%% file: main_document.tex
\documentclass[journal]{new-aiaa}

\input{macros}

\makeatletter
\patchcmd{\ttlh@hang}{\parindent\z@}{\parindent\z@\leavevmode}{}{}
\patchcmd{\ttlh@hang}{\noindent}{}{}{}
\makeatother

\title{Verification and Validation of Convex Optimization Algorithms for Model Predictive Control}

\author{Raphael Cohen\footnote{Ph.D. Student, School of Aerospace Engineering, raphael.cohen@gatech.edu.} and Eric Feron \footnote{Professor of Aerospace Engineering, School of Aerospace Engineering, feron@gatech.edu.} }
\affil{Georgia Institute of Technology, School of Aerospace Engineering, Atlanta, GA 30332}
\author{Pierre-Lo\"{i}c Garoche\footnote{Research Scientist, Onera, pierre-loic.garoche@onera.fr.}}
\affil{Onera -- The French Aerospace Lab, Toulouse, 31000 France}
\begin{document}
\maketitle
\begin{abstract}	
Advanced embedded algorithms are growing in complexity and they are 
an essential contributor to the growth of autonomy in many areas. 
However, the promise held by these algorithms cannot be kept without 
proper attention to the considerably stronger design constraints that 
arise when the applications of interest, such as aerospace systems, are 
safety-critical. Formal verification is the process of proving or 
disproving the "correctness" of an algorithm with respect to a certain 
mathematical description of it by means of a computer. 
This article discusses the formal verification of the Ellipsoid method, 
a convex optimization algorithm, and its code implementation as it applies
to receding horizon control. Options for encoding code properties and their proofs
are detailed. The applicability and limitations of those code properties
and proofs are presented as well.
Finally, floating-point errors are taken into account in a numerical analysis 
of the Ellipsoid algorithm. Modifications to the algorithm are presented which 
can be used to control its numerical stability.
\end{abstract}

\section*{Nomenclature}
{\renewcommand\arraystretch{1.0}
\noindent\begin{longtable*}{@{}l @{\quad=\quad} l@{}}
$\mathbb{R}$ 				& The set of all real numbers. \\
$\mathbb{F}$				& Set of floating-point numbers. \\
$\mathbb{R^{+}}$ 			& The set of all positive real numbers. \\
$\mathbb{R^{*}_{+}}$		& The set of all strictly positive real numbers. \\
$\mathbb{R}^n$				& The set of real vectors of length $n$.  \\
$\mathbb{R}^{m \times n}$	& The set of real matrices of size $m \times n$.  \\
$\norm{A}_{F}$ 				& Frobenius norm of a matrix $A$. \\
$\norm{A}$     				& Two norm of a matrix $A$. \\
$\norm{x}$     				& Two norm of a vector $x$. \\
$B_{n}$						& n-dimensional unit Euclidean ball. 
		$B_{n} = \{ z \in \mathbb{R}^{n} : \norm{z} \leq 1  \}$.  \\
$B_{r}(x)$					& Ball of radius $r$ centered on $x$. 
		$B_{r}(x) = \{ z \in \mathbb{R}^{n}: \norm{z-x} \leq r  \}$.  \\
$\Ell(B,c)$					& Ellipsoid set defined by: $\Ell(B,c) = \{ Bu+c : u \in B_n \}$.  \\	
$\fl()$						& Floating-point rounding of a given real number. \\
$H$ 						& Model Predictive Control Horizon.  \\
$k(A)$ 						& Condition number of a matrix $A$. \\
$N$							& Number of iterations.  \\
$u$							& Plant Input. \\
$\textbf{u}$ 				& Collection of input vectors to horizon:
		$\textbf{u} = [u_1 \dots u_{H-1}]$.  \\

$\Vol()$					& Volume of a given set.  \\
$X$ 						& Original Decision Vector for an Optimization problem. \\
$x$							& Plant State Vector. \\
$\textbf{x}$ 				& Collection of state vectors to horizon: 
		$\textbf{x} = [x_1 \dots x_H]$.  \\
$X_f$						& Feasible set of an optimization problem. \\
$X_{\epsilon}$				& Epsilon optimal set of an optimization problem. \\
$Z$ 						& Projected Decision Vector for an Optimization problem. \\

$\gamma$					& Upper bound of Reduction Ratio. \\
$\theta$					& Elevation Angle in Radians for the three degrees of freedom Helicopter. \\
$\lambda$					& Ellipsoids Widening Coefficient. \\
$\sigma_{\rm max}(A)$ 		& Largest singular value of a matrix $A$. \\
$\sigma_{\rm min}(A)$ 		& Smallest singular value of a matrix $A$. \\
$\phi$						& Travel Angle in Radians for the three degrees of freedom Helicopter. \\
$\psi$						& Pitch Angle in Radians for the three degrees of freedom Helicopter. \\
\end{longtable*}}


\input{intro}

\input{prelim}
\input{cvxEll}
\input{online}
\input{correction}
\input{flp}
\input{experiment}
\input{conclusion}
\section*{Acknowledgments}
\noindent This work was partially supported by projects
ANR ASTRID VORACE, ANR FEANICSES ANR-17-CE25-0018, and NSF CPS 
SORTIES under grant 1446758. The authors would also like to thank 
Pierre Roux from ONERA for reviewing this work, and Arkadi Nemirovski 
for sharing how the Ellipsoid method can be modified to keep the
ellipsoid's condition number bounded.

\setlength{\bibsep}{0pt plus 0.3ex}

\setstretch{1.0}
\bibliography{biblio}

\end{document}

%% file: macros.tex
\usepackage[utf8]{inputenc}
\usepackage[version=4]{mhchem}
\usepackage{siunitx}
\usepackage{longtable,tabularx}
\usepackage[margin=1in]{geometry} 
\usepackage[english]{babel}
\usepackage{setspace}
\usepackage{wrapfig}
\usepackage{hyperref}
\usepackage{amsmath,amsfonts,amssymb}
\usepackage{stmaryrd} 
\usepackage{xspace}
\usepackage[normalem]{ulem}
\usepackage[dvipsnames]{xcolor}
\usepackage{listings}
\usepackage{graphicx}
\usepackage{comment}
\usepackage{url}
\usepackage{bussproofs}
\usepackage[noend]{algorithmic}
\usepackage{algorithm}
\usepackage{cancel}
\usepackage{mathtools}
\usepackage{moreverb}
\usepackage{dirtytalk}
\usepackage{subcaption}

\hypersetup{
    colorlinks,
    linkcolor={red!50!black},
    citecolor={blue!50!black},
    urlcolor={blue!80!black}
}

\DeclarePairedDelimiter\abs{\lvert}{\rvert}

\newtheorem{mylemma}{Lemma}
\newtheorem{myfact}{Fact}
\newtheorem{mydef}{Definition}
\newtheorem{myprop}{Property}
\newtheorem{myth}{Theorem}

\newtheorem{po}{Proof}

\newcommand{\Vol}{\text{{\normalfont Vol}}}
\newcommand\Ell{\text{{\normalfont Ell}}}
\newcommand\fl{\text{{\normalfont fl}}}
\newcommand\norm[1]{\left\lVert#1\right\rVert}
\newcommand*{\QEDB}{\hfill\ensuremath{\square}}%
\definecolor{mGreen}{rgb}{0,0.6,0}
\definecolor{mGray}{rgb}{0.5,0.5,0.5}
\definecolor{mPurple}{rgb}{0.58,0,0.82}
\definecolor{backgroundColour}{rgb}{0.95,0.95,0.92}
\definecolor{OliveGreen}{rgb}{0,0.6,0}

\lstdefinestyle{cacsl}{
	language=C++,
    commentstyle=\color{mGreen},
    keywordstyle=\color{blue},
    numberstyle=\tiny\color{mGray},
    stringstyle=\color{mPurple},
    alsoletter={\\},
    keywords = [2]{	complete, behaviors, disjoint, loop, invariant, 
    				requires, ensures, predicate, axiom, \\nothing, 
    				\\exists, \\result, assigns, \\forall , type, 
    				axiomatic, logic, reads, real, integer, assumes,
    				behavior},
  	keywordstyle=[2]\color{red},
  	morecomment=[s][\color{OliveGreen}]{/*}{*/},
  	moredelim=**[s][\color{OliveGreen}]{/*@}{*/},
  	moredelim=**[l][\color{OliveGreen}]{//@},
    basicstyle=\linespread{0.4}\fontsize{8}{12}\ttfamily,
    breakatwhitespace=false,         
    breaklines=false,                 
    captionpos=b,                    
    keepspaces=true,                 
    numbers=left,                    
    numbersep=5pt,                  
    showspaces=false,                
    showstringspaces=false,
    showtabs=false, 
    xrightmargin=0.1cm,          
    tabsize=2
}

\definecolor{light-gray}{gray}{0.90}

\lstdefinestyle{inputFile}{
	language=C++,
    commentstyle=\color{mGreen},
    keywordstyle=\color{blue},
    numberstyle=\tiny\color{mGray},
    stringstyle=\color{mPurple},
    keywords = [2]{Information, Input, Output, Variables, Constants , Minimize , SubjectTo},
  	keywordstyle=[2]\color{red},
  	morecomment=**[l][\color{OliveGreen}]{\%},
    basicstyle=\linespread{0.5}\fontsize{8}{12}\ttfamily,
    breakatwhitespace=false,         
    breaklines=false,                 
    captionpos=b,                    
    keepspaces=true,                 
    numbers=left,                    
    numbersep=5pt,                  
    showspaces=false,                
    showstringspaces=false,
    showtabs=false, 
    xrightmargin=0.1cm,          
    tabsize=2,
}

\usepackage{tcolorbox}
\tcbuselibrary{listings,skins}

\newtcblisting{cacslListing}[2][]{
    arc=1pt, outer arc=5pt,
    listing only, 
    listing style=cacsl,
    title=#2,#1
}

\newtcblisting{genemoListing}[2][]{
    arc=1pt, outer arc=5pt,
    listing only,
    hbox, 			
    listing style=inputFile,
    title=#2,#1
}

%% file: intro.tex
\section{Introduction}
\label{sec:intro} 
\noindent Formal verification of optimization algorithms used online within 
control systems is the sole focus of this research. Recently, such algorithms have
been used online with great success for the guidance of safety-critical applications,
including, autonomous cars~\cite{DBLP:journals/tac/JerezG0CKM14} and reusable 
rockets~\cite{DBLP:journals/tcst/AcikmeseCB13}. 
The latter case has resulted in technology demonstrations such as the landings of SpaceX's 
Falcon 9~\cite{2016bridge} and BlueOrigin's New Shepard. Thus, algorithms solving optimization 
problems are already used online, have been embedded on board, and yet 
still lack the level of qualification required by civil aircraft or manned rocket flight.
Automatic code generation for solving convex optimization problems has already been done \cite{boyd2004convex,mattingley2012cvxgen},
but does not include the use of formal methods. Likewise, work within the field of model predictive 
control already exists where numerical properties of algorithms are being evaluated 
\cite{patrinos2015dual}. Nevertheless, this work is only valid for Quadratic Programming 
(See Section~\ref{sec:prelim}) and using fixed-point arithmetic. As well, no formal verification is performed.
On the other hand, some contributions have been made concerning formal verification 
of control systems~\cite{Feron2010,Champion2013,Heber2012,Roux2012Ell}, but they
mainly focus on formal verification and code generation for linear control systems. 
Research has also been made toward the verification 
of numerical optimization algorithms~\cite{Wang2016,wei2006numerical}, yet it remains purely theoretical and 
no proof was obtained using formal verification tools. Contributions on formal verification of optimization 
algorithms have already been made~\cite{cohen2017formal}, but this work focuses on a single optimization 
problem, where closed-loop behaviors are not being addressed, which does not meet the level of guarantees 
needed for receding horizon controllers.
As well, the formal proof was not complete and no numerical analysis was presented. \\
The need for enhanced safety and better performance is currently 
pushing for the introduction of advanced numerical methods into the next generation of 
cyber-physical systems. While most of the algorithms described in this article have been 
established for a long time, their online use within embedded systems is relatively 
new and introduces issues that have to be addressed. Among these methods, this study focuses
on numerical optimization algorithms. \\
The following scientific contributions are presented:
\begin{itemize}
\item axiomatization of optimization problems and formalization 
of algorithm proof (Ellipsoid method) as code annotation
\item extraction of guarantees of convergence for sequential optimization problems, 
	representing closed-loop management
\item modification of the original algorithm to account for floating-point errors 
\item generation of C code implementations via credible autocoders of receding horizon 
controllers along with ANSI/ISO C Specification Language (ACSL) annotations
\end{itemize}
The choice of ellipsoid method here seems unconventional as current state of art 
solvers typically use some variant of the interior-point method. However it has 
been shown in~\cite{cohen2017formal} that guaranteeing the numerical accuracy of second-order 
methods are very challenging.
This paper is a first attempt at providing methods and tools to formally verify 
convex optimization code for solving online receding-horizon control problems.
The article is structured as follows: Section~\ref{sec:prelim} presents
backgrounds for convex optimization, model predictive control, and axiomatic semantics using Hoare
triples. Section~\ref{sec:cvxEll} focuses on the axiomatization of 
second-order cone programs and the formal verification of the Ellipsoid Method. 
Furthermore, the closed-loop management and the online aspect of the developed algorithm 
is discussed in Section~\ref{sec:online}. A modified version of the original 
Ellipsoid Method used to control floating-point errors, is presented in 
Section~\ref{sec:correction}. A floating-point analysis of the ellipsoid method 
is presented in Section~\ref{sec:flp}, while Section~\ref{sec:experiment} presents 
how this framework can be automated and applied to a system, the three degrees of freedom (DOF) Helicopter. 
Finally, Section~\ref{sec:conclusion} concludes this article.

%% file: prelim.tex
\section{Preliminaries}
\label{sec:prelim}
\subsection{Second-Order Cone Programming}
\label{subsec:cvxopt}
\noindent Optimization algorithms solve a constrained optimization problem, defined by an objective
function and a set of constraints to be satisfied:
\vspace{-10pt}
\begin{equation}
\label{eq:opt_pb}
\begin{split}
\min\qquad & f_o (x) \\
\text{s.t.}\qquad & f_i(x) \leq b_i \text{ for } i \in [1,m]
\vspace{-30pt}
\end{split}
\end{equation}
\noindent This problem searches for $x \in \mathbb{R}^{n}$, the optimization variable, minimizing 
$f_{o} \in \mathbb{R}^{n} \rightarrow \mathbb{R}$, the objective function, while satisfying 
constraints $f_{i} \in \mathbb{R}^{n} \rightarrow \mathbb{R}$, with associated bounds
$b_{i}$. An element of $\mathbb{R}^{n}$ is feasible when it satisfies all the constraints 
$f_{i}$. An optimal point is defined by the element having the smallest cost value among 
all feasible points. An optimization algorithm computes an exact or approximated estimate 
of the optimal cost value, together with one or more feasible points achieving this 
value. A subclass of these problems that can be efficiently solved are convex optimization 
problems. In these cases, the functions $f_{o}$ and $f_{i}$ are required to be convex~\cite{boyd2004convex},
with one of the consequences being that a local minimizer is also a global minimizer. 
Furthermore, when the constraints are linear the problem is either called a Linear Program (LP)
if the cost is also linear or called a Quadratic Program (QP) if the cost is quadratic.
Optimization problems where both the cost and the constraints are quadratic are called 
Quadratically Constrained Quadratic Program (QCQP).
Semi-Definite Programs (SDP) represent problems that have constraints which can be formulated 
as a Linear Matrix Inequality (LMI).
Here, only a specific subset of convex optimization problems are presented in details: Second-Order Cone 
Programs (SOCPs). For $x \in \mathbb{R}^{n}$, a SOCP in standard form can be written as:
\begin{equation}
\begin{split}
\min\qquad & f^{T} x \\
\text{s.t.}\qquad & \norm{A_{i} x + b_{i}}_{2} ~ \leq ~ c_{i}^{T} x + d_{i} \text{ for } i \in [1~,~m] \\
\text{With: } f \in \mathbb{R}^{n}, ~ & A_{i} \in \mathbb{R}^{n_{i} \times n},
~b_{i} \in \mathbb{R}^{n_{i}}, ~ c_{i} \in \mathbb{R}^{n}, ~ d_{i} \in \mathbb{R}.
\label{eq:socp_pb}
\end{split}
\end{equation}
\begin{wrapfigure}{r}{0.4\textwidth} 
\vspace{-30pt}
  \begin{center}
    \includegraphics[width=0.4\textwidth]{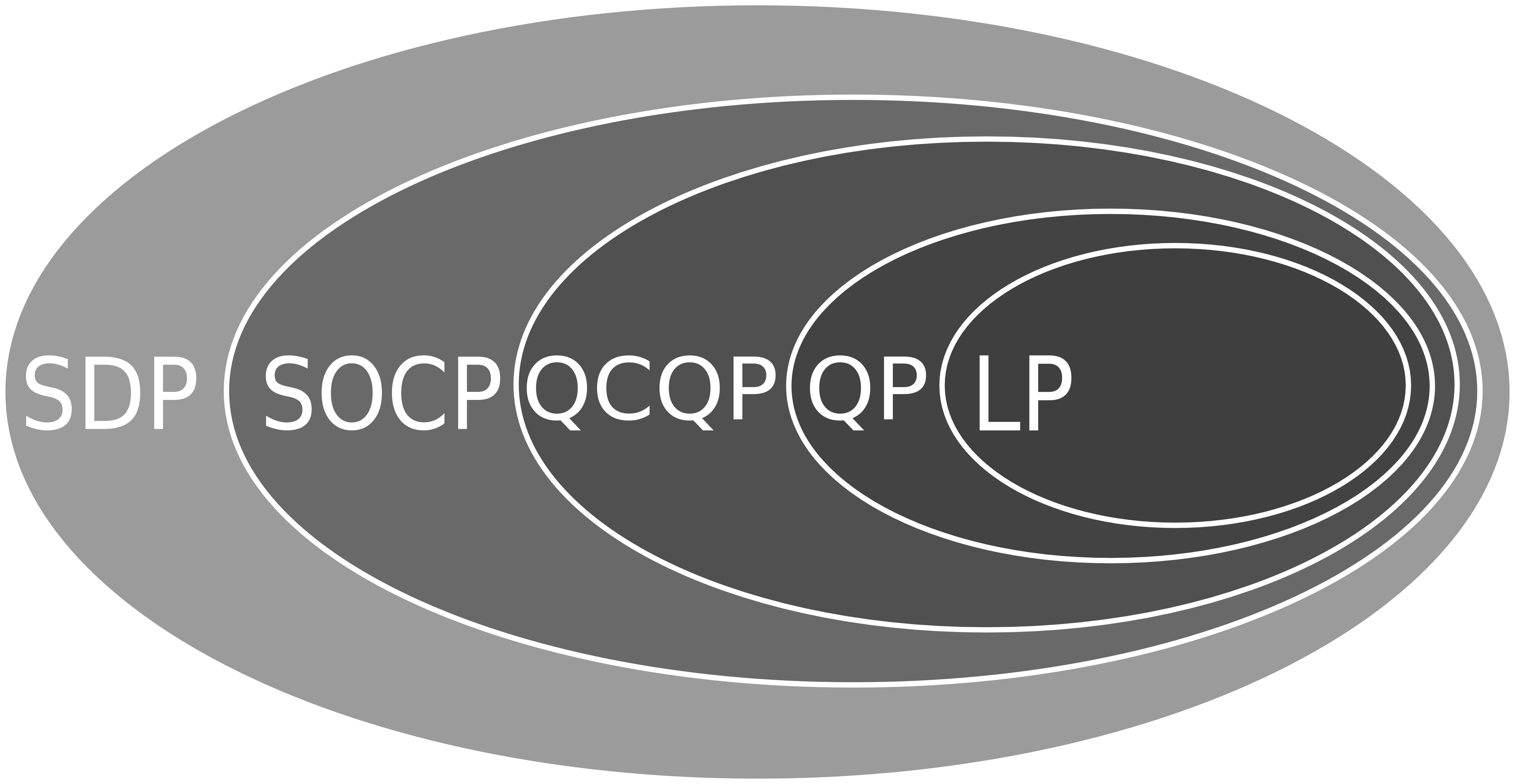}
    \caption{Classification of Some Convex Optimization Problems}
    \label{fig:classification}
  \end{center}
  \vspace{-40pt}
\end{wrapfigure} 
A classification of the most common convex optimization problems is presented in
Fig.~\ref{fig:classification}. Frequently, optimization problems that are used online 
for control systems can be formulated as a SOCP. Furthermore, extensions to SDPs are 
possible with little additional work. The algorithm used and the proof are still valid for any convex 
problem.
\subsection{Model Predictive Control (MPC)}
\noindent Model predictive control (also known as receding horizon control) is an optimal control 
strategy based on numerical optimization. In this technique, a discrete-time dynamical 
model of the system is being used to predict potential future trajectories. As well, a cost 
function, that depends on the future control inputs and states, is being considered over 
the receding prediction horizon $H$, with the objective being to minimize this cost.
At each time $t$, a convex optimization problem is being solved and the corresponding input is sent to the system. 
A time step later, the exact same process occurs and is repeated until a final time. We refer the reader 
to~\cite{blackmore2010minimum,borrelli2005mpc} for more details on MPC. 
\vspace{-5pt}
\subsection{Axiomatic Semantics and Hoare Logic}
\noindent Semantics of programs express their behavior. Using axiomatic semantics,
the program's semantics can be defined in an incomplete way, as a set of projective statements, 
i.e., observations. This idea was formalized by~\cite{Floyd1967Flowcharts} and 
then~\cite{hoareaxiom69} as a way to specify the expected behavior of a program through 
pre- and post-conditions.
\vspace{-15pt}
\paragraph*{Hoare Logic.}
A piece of code $C$ is axiomatically described by a pair of formulas $(P,Q)$ such
that when $P$ holds before executing $C$, then $Q$ should be valid after its
execution. This pair acts as a contract for the function and $(P,C,Q)$ is called
a Hoare triple. In most uses, $P$ and $Q$ are expressed in first order formulas
over the variables of the program. Depending on the level of precision of these
annotations, the behavior can be fully or partially specified.  In our case we
are interested in specifying, at code level, algorithm specific properties such
as the algorithm convergence or preservation of feasibility for
intermediate iterates. 
Software analyzers, such as the Frama-C platform~\cite{frama-c},
provide means to annotate source code with these contracts, and tools to
reason about these formal specifications.
For the C language, ACSL~\cite{acsl} 
(ANSI C Specification Language) can be used to write source comments. \\
\vspace{-10pt}
\begin{wrapfigure}{r}{0.38\textwidth} 
\vspace{-15pt}
\begin{center}
\begin{cacslListing}[hbox,enhanced,drop shadow]{ACSL + C}
/*@ 
  @ requires -2 <= x <= 2;
  @ ensures \result == x*x;
  @ ensures 0 <= \result <= 4;
  @ assigns \nothing;
*/
double square(double x){
	return x*x;
}
\end{cacslListing}
\caption{ACSL Function Contract}
\label{fig:fct_contract}
  \end{center}
  \vspace{-20pt}
  \vspace{1pt}
\end{wrapfigure} 
%
Figure~\ref{fig:fct_contract} shows an example of a function contract
expressed in ACSL. The \say{ensures} keyword expresses all the properties 
that will be true after the execution of the function, assuming that all the properties listed
within the \say{requires} keywords were true before the execution (similar to a Hoare triple). 
As well, it is possible to annotate and check the part of the memory assigned by a 
function using the keyword \say{assigns}. In the case shown in 
Fig.~\ref{fig:fct_contract}, the function is not assigning anything during its execution, 
and therefore no global variable were changed. Throughout this article, the verification 
is performed using the software analyzer Frama-C and the SMT solver (Satisfiability 
Modulo Theories) Alt-Ergo~\cite{alt-ergo}, via the Weakest Precondition (WP) plug-in.
The role of the WP plug-in is to implement a weakest precondition calculus for ACSL 
annotations present at code level. For each annotation, the WP plug-in generates
proof obligations (mathematical first-order logic formulas) that are then submitted to 
Alt-Ergo. Further information about the WP plug-in can be found in~\cite{frama-c-wp}.

%% file: cvxEll.tex
\section{Formal Verification of an Ellipsoid Method C code Implementation}
\label{sec:cvxEll}
\noindent Our goal is to build a framework that is capable of compiling the 
high-level requirements of online MPC solvers into ACSL augmented C code
which can then be automatically verified using existing formal methods 
tools for C programs such as Frama-C and Alt-ergo.
The MPC solver shall take a parameterized SOCP problem as inputs and 
always outputs a solution that is both feasible and epsilon-optimal 
within a predefined number of iterations. 
\newline \noindent This kind of requirement has never been formalized before. 
Hence it also has never been verified by the state of art automatic 
formal methods techniques and it is not really possible to do so 
without going into some manual proofs. To formalize these high-level 
requirements, one has to:
\begin{itemize}
\item formalize in ACSL the low-level types, such as vectors and matrices
\item formalize in ACSL second-order cone problems 
\item formalize the solver (from algorithm and input problem to 
the generated C code)
\item formalize the properties of the ellipsoid method in a way such 
that they can be expressed as axiomatic semantic of the C program 
(ACSL types, axioms, functions to express feasability and optimality).
\end{itemize}
For example, mathematical types from linear algebra can be defined 
axiomatically. The input problem and solver choice is automatically 
transformed into a C program. The high-level properties (feasability and 
optimality) of the chosen solver together with the input problem are 
compiled into an ACSL form (expressing the axiomatics semantics of 
the generated C program), and then inserted into the C program as comments.  
The various artifacts (ACSL types, functions, predicates, axioms, lemmas, 
theorems), that were manually written to support the compilation of the 
high-level requirements (HLR) into C+ACSL and its automatic verification, 
are packaged into libraries. Examples of these artifacts 
include types like matrix, vector, optim, predicates like 
\say{isFeasible} and functions such as \say{twoNorm}, etc.
\subsection{Semantics of an Optimization Problem}
\label{sec:semOpt}
\noindent The first work to be done is the formal definition of an optimization problem. 
In order to do so, new mathematical types, objects, axioms and theorems are created.
Our goal is to axiomatize optimization problems with enough properties allowing 
us to state all the needed optimization-level properties at code level.
Let us consider the second-order cone program, described in Eq.~\eqref{eq:socp_pb}. \\
\emph{Encoding an SOCP.} In order to fully describe an SOCP, we use the variables: \\
\[ f \in \mathbb{R}^{n}, ~~ A = 
\begin{bmatrix} 
A_{1} \\[-5pt]
\vdots\\[-5pt]
A_{m} \\ 
\end{bmatrix} , ~~  b = 
\begin{bmatrix} 
b_{1} \\[-5pt]
\vdots\\[-5pt]
b_{m} \\ 
\end{bmatrix} , ~~ C = 
\begin{bmatrix} 
c_{1}^{T} \\[-5pt]
\vdots\\[-5pt]
c_{m}^{T} \\ 
\end{bmatrix}, ~~ d = 
\begin{bmatrix} 
d_{1} \\[-5pt]
\vdots\\[-5pt]
d_{m} \\ 
\end{bmatrix} ~~ \text{and also the vector}  ~~ m = \begin{bmatrix} 
n_{1} &  \dots & n_{m} \end{bmatrix} . \]
The vector $m$ is used to collect the sizes of the vectors
$A_{i}\cdot x + b_{i}$. Furthermore, 
if $\sum_{i=1}^{m} n_{i}=0$, then the SOCP~\eqref{eq:socp_pb} is
an LP. Using ACSL, a new type and a high level function are defined, 
providing the possibility to create objects of the type \say{$optim$}. 
Figure~\ref{fig:SocpAxiom} represents an extract of the ACSL optimization theory.
First, a new ACSL theory is created using the keyword \say{axiomatic}. 
The keyword \say{logic} is used to define a new function
and its signature. Information about the part of memory used by a function is 
provided using the keyword \say{reads}. Figure~\ref{fig:SocpAxiom} presents the definition 
of 2 functions. The function $socp\_of\_size\_2\_6\_0$ is used to instantiate objects 
representing an optimization problem of appropriate sizes.
\begin{wrapfigure}{r}{0.45\textwidth} 
\vspace{-15pt}
  \begin{center}
\begin{cacslListing}[hbox,enhanced,drop shadow]{ACSL}
/*@ 
axiomatic OptimSOCP {
type optim;
logic optim socp_of_size_2_6_0(
   matrix A,vector b,matrix C, 
   vector d, vector f, int* m) 
   reads m[0..5];
logic vector constraints(optim OPT,
   vector x);	
*/
\end{cacslListing}
\caption{ACSL Optim Type Definition}
\label{fig:SocpAxiom}
  \end{center}
  \vspace{-10pt}
\end{wrapfigure} 
The function $constraints$ returns
a vector collecting the values of all the 
constraint functions for a given problem and point.
When applying a method to solve an actual optimization 
problem, many concepts are crucial. 
The work here is to highlight the parts of the HLR that need to 
be formalized (via axiomatization) and packaged into a 
library to support the automatic compilation of the HLR into ACSL augmented C code.
The concepts of feasibility and optimality are being axiomatized. For this, given 
a second-order cone program, an axiomatic definition is given for the vector 
constraint, the gradient of a constraint, the cost, optimal point (making the 
assumption that it exists and is unique), etc.
For instance, Fig.~\ref{fig:FeasiblePredicate} illustrates the axiomatization 
of a constraint calculation and the feasibility predicate definition.
For the constraint calculation, two axioms are defined representing two different
cases: The case where the constraint is linear and the case where it is not.
The predicate shown in Fig.~\ref{fig:FeasiblePredicate} 
defines that a point is feasible if all the components of its constraint vector are negative.
\begin{figure}[ht!]
\centering
\begin{cacslListing}[hbox,enhanced,drop shadow]{ACSL}
/*@
  axiom constraint_linear_axiom:
    \forall optim OPT, vector x, integer i;
      getm(OPT)[i] == 0 ==>	
        constraint(OPT, x, i) == 
          -scalarProduct(getci(OPT,i),x,size_n(OPT))-getdi(OPT,i);
  axiom constraint_socp_axiom:
    \forall optim OPT, vector x, integer i;
      getm(OPT)[i] != 0 ==>	
        constraint(OPT, x, i) == 
          twoNorm(vector_affine(getAi(OPT,i),x,getbi(OPT,i))) -
          scalarProduct(getci(OPT,i),x,size_n(OPT))-getdi(OPT,i);
  ...
  predicate
    isFeasible(optim OPT,vector x) = isNegative(constraints(OPT,x));
*/
\end{cacslListing}
\caption{ACSL Feasible Predicate Definition}
\label{fig:FeasiblePredicate}
\end{figure}
%
When instantiating an object of type vector or matrix, the size of the considered object
needs to be known since it is hard-coded in the ACSL axiomatization.
This is not an issue since at this time (post parsing), all the 
sizes of the variables used are already defined. 
The objects sizes only depend on the plant's order and the horizon. Thus,
as long as the order of the plant and the horizon do not change dynamically, 
the variables sizes can be predicted.
Also, working with predefined and hard-coded size objects will help the analyzers proving the
goals.

%
\vspace{-10pt}
\subsection{The Ellipsoid Method}
\label{sec:EllMethod}
\begin{figure}[ht!]
  \centering
  \begin{subfigure}{.5\textwidth}
  \centering
  \includegraphics[width=0.50\linewidth]{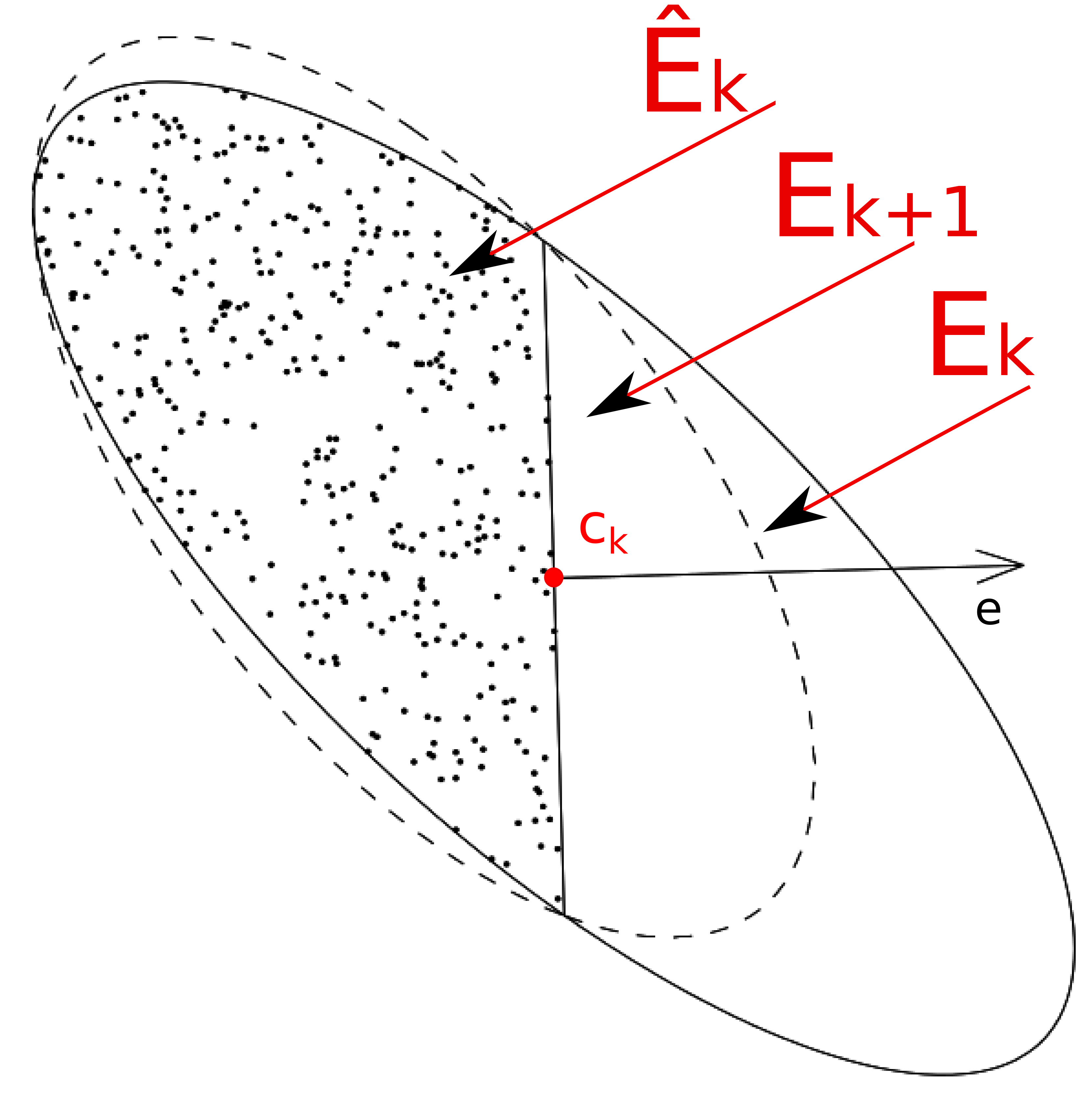}
  \caption{Ellipsoid Cut}
  \label{fig:ellipsoid_cut}
  \end{subfigure}%
  \begin{subfigure}{.5\textwidth}
  \centering
  \includegraphics[width=0.61\linewidth]{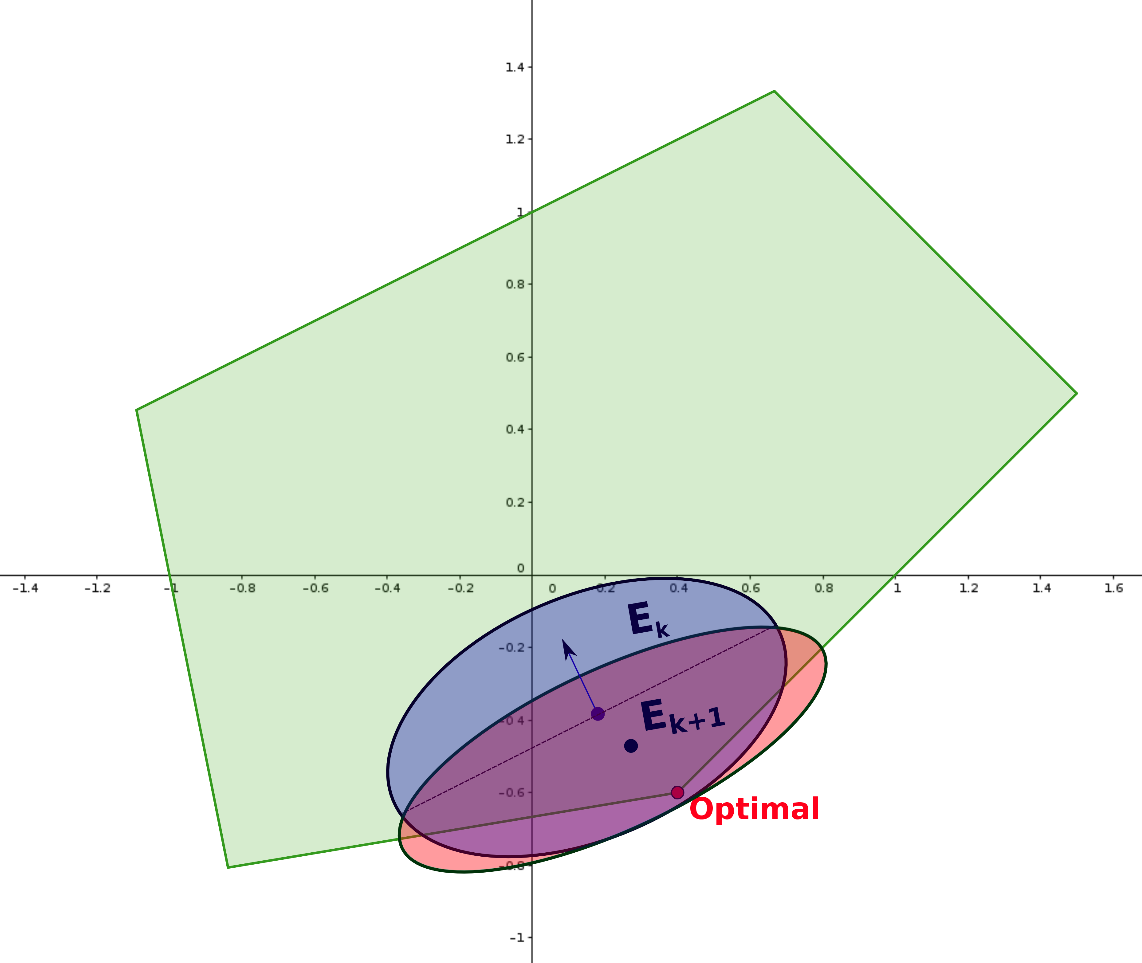}
  \caption{Ellipsoid Cut In a LP Setting}
  \label{fig:ellipsoid_cut_lp}
  \end{subfigure}
\caption{Ellipsoids Cut Illustration}
\end{figure}

\noindent Despite its modest efficiency with respect to interior point methods,
the Ellipsoid Method~\cite{grotschel1981ellipsoid,bland1981ellipsoid,Nemirovski2012}
benefits from concrete proof elements and could be
considered a viable option for critical embedded systems where safety is more
important than performance. This section presents a way to annotate 
a C code implementation of the Ellipsoid Method. Before recalling the main 
steps of the algorithm, some mathematical preliminaries are presented.
%
%
\vspace{-10pt}
\paragraph*{Ellipsoids in $\mathbb{R}^{n}$.}
An ellipsoid can be characterized as an affine transformation of an Euclidean Ball. 
Before defining an Ellipsoid set, the definition of an Euclidean ball is first recalled.
\begin{mydef}[Euclidean ball]
Let $n \in \mathbb{N}$, $V_{n}$ denotes the unit Euclidean ball in
$\mathbb{R}^{n}$. $\Vol(V_{n})$ represents its volume.
As well, $B_{R}(x)$ is defined as the ball of radius $R$ centered on $x$
\Big( i.e $\{ z \in \mathbb{R}^{n}: (z-x)^{T}(z-x) \leq R \}$ \Big).
\end{mydef}
\begin{mydef}[Ellipsoid Sets]
Let $c \in \mathbb{R}^{n}$ and $B \in \mathbb{R}^{n \times n}$, be a non-singular 
matrix ($det(B) \neq 0$). The Ellipsoid $\Ell(B,c)$ is the set :
\vspace{-5pt}
\begin{equation}
\Ell(B,c) = \{ Bu+c : u^{T}u \leq 1 \}
\label{eq:ellipsoid_def}
\end{equation} 
\end{mydef}
\begin{mydef}[Volume of Ellipsoids]
Let $\Ell(B,c)$ be an ellipsoid set in $\mathbb{R}^{n}$. $Vol(\Ell(B, c))$ denotes its
volume and is defined as :
\vspace{-5pt}
\begin{equation}
\Vol(\Ell(B,c)) = |det(B)| \cdot \Vol(V_{n})
\label{eq:volume_ellipsoid}
\end{equation}  
\end{mydef}
\paragraph*{Algorithm.} The main steps of the algorithm detailed 
in~\cite{bland1981ellipsoid,Nemirovski2012,boyd1991linear} are now presented.
In the following, $E_{k} = \Ell (B_k,c_k)$ denotes the ellipsoid computed 
by the algorithm at the $k^{th}$ iteration.
\paragraph*{Ellipsoid cut.}The algorithm starts with an ellipsoid containing the feasible set $X$,
and therefore the optimal point $x^{*}$. An iteration consists of transforming the current
ellipsoid $E_{k}$ into a smaller volume ellipsoid $E_{k+1}$ that also contains
$x^{*}$. Given an ellipsoid $E_{k}$ of center $c_{k}$, the objective is to find a
hyperplane containing $c_{k}$ that cuts $E_{k}$ in half, such that one half
is known not to contain $x^{*}$. Finding such a hyperplane is called the \textit{oracle separation}
step, \textit{cf.}~\cite{Nemirovski2012}. Within the SOCP setting, this cutting hyperplane is 
obtained by taking the gradient of either a violated constraint or the cost function. 
Then, the ellipsoid $E_{k+1}$ is defined by the
minimal volume ellipsoid containing the half ellipsoid $\hat{E}_{k}$ that is
known to contain $x^{*}$. The Fig.~\ref{fig:ellipsoid_cut} and~\ref{fig:ellipsoid_cut_lp} 
illustrate such ellipsoids cuts.
\paragraph*{Ellipsoid transformation.}
From the oracle separation step, a separating hyperplane, $e$, that cuts $E_{k}$ in half 
with the guarantee that $x^{*}$ is localized in $\hat{E}_{k}$ has been computed. The following 
step is the \textit{Ellipsoid transformation}. Using this hyperplane $e$, one can update 
the ellipsoid $E_{k}$ to its next iterate $E_{k+1}$  according to 
Eqs.~\eqref{eq:updateCenter},\eqref{eq:updateShape} and \eqref{eq:updateHyperplane}. In addition to that,
an upper bound, $\gamma$, of the ratio of $\Vol(E_{k+1})$ to $\Vol(E_{k})$ is known.
\vspace{-0pt}
\begin{equation}
c_{k+1} = c_{k} - 1/ (n+1)\cdot  B_{k} p ~~, 
\label{eq:updateCenter}
\end{equation}  
\vspace{-15pt}
\begin{equation}
B_{k+1} = \frac{n}{\sqrt{n^2 - 1}} B_{k} +  \bigg(  \frac{n}{n+1} - \frac{n}{\sqrt{n^2 -1}}   \bigg) (B_{k}p)p^{T}
\label{eq:updateShape}
\end{equation}  
with: 
\vspace{-10pt}
\begin{equation}
p = B_{k}^{T}e / \sqrt{e^{T}B_{k}B_{k}^{T}e} .
\label{eq:updateHyperplane}
\end{equation}
\vspace{-15pt}
\paragraph*{Termination.}
The search points are the successive centers of the ellipsoids. Throughout the 
execution of the algorithm, the best point so far, $\hat{x}$ is being stored in memory.
A point $x$ is better than a point $y$ if it is feasible and has a smaller cost. 
When the program reaches the number of iteration needed, the best point so far, 
$\hat{x}$, which is known to be feasible and $\epsilon$-optimal, is returned by 
the algorithm. A volume related property is now stated, at the origin of the 
algorithm convergence, followed by the main theorem of the method. Both properties
can be found in~\cite{Nemirovski2012,khachiyan1980polynomial}.

\begin{myprop}{[Reduction ratio.]}
Let $k \geq 0$, by construction:
\vspace{-5pt}
\begin{equation}
\Vol(E_{k+1}) \leq \exp  \{ -1 / (2 \cdot (n+1))  \}  \cdot \Vol(E_{k})
\label{eq:volume_ratio}
\end{equation}  
\label{prop:volume_ratio}
\end{myprop}
\vspace{-25pt}
Please find below the the proof of this property.
\begin{po}
Let us put the update formula~\eqref{eq:updateShape} into the form:
\vspace{-5pt}
\[ 
\vspace{-5pt}
B_{k+1} = \alpha B_{k} + \beta (B_{k}p)p^{T} \]
With: $\alpha = n/\sqrt{n^2 -1}$ and $\beta = n/(n+1) - n/\sqrt{n^2 -1}$.
Let us now take the determinant of both sides.
\vspace{-5pt}
\begin{align}
\det (B_{k+1})   & = \det \Big( B_{k} \cdot \big( \alpha I_{n} + \beta pp^{T} \big) \Big) = \det \big( B_{k} \big) \det \Big(  \alpha I_{n} + \beta pp^{T}  \Big) = \det \big( B_{k} \big) \alpha^{n} \det \Big(  I_{n} + \frac{\beta}{\alpha} pp^{T} \Big) \nonumber
\end{align}
Using Sylvester's determinant identity:
$ \det (I_n + AB) = \det(I_m + BA) ~~~~ \forall A \in \mathbb{R}^{n \times m}, B \in \mathbb{R}^{m \times n} $,
the determinant on the right side of the equality can be put into the form:
\vspace{-5pt}
\[ \vspace{-5pt}
\det (B_{k+1})  =  \alpha^{n} \det \big( B_{k} \big) \cdot \Big(  1 + \frac{\beta}{\alpha} \norm{p} \Big)  \]
But, from Eg.~\eqref{eq:updateHyperplane}, one can see that $\norm{p} = 1$. Therefore:
\[ \frac{\Vol(E_{k+1})}{\Vol(E_{k})} =  \frac{ | \det (B_{k+1})| }{ | \det (B_{k}) | }  =  \alpha^{n} \cdot 
\Big(  1 + \frac{\beta}{\alpha}  \Big)  \leq \exp \Big( \frac{-1}{2(n+1)} \Big) \]
\QEDB
\end{po}
%
%
\begin{wrapfigure}{r}{0.45\textwidth} 
\vspace{-30pt}
\begin{center}
\includegraphics[width=0.35\textwidth]{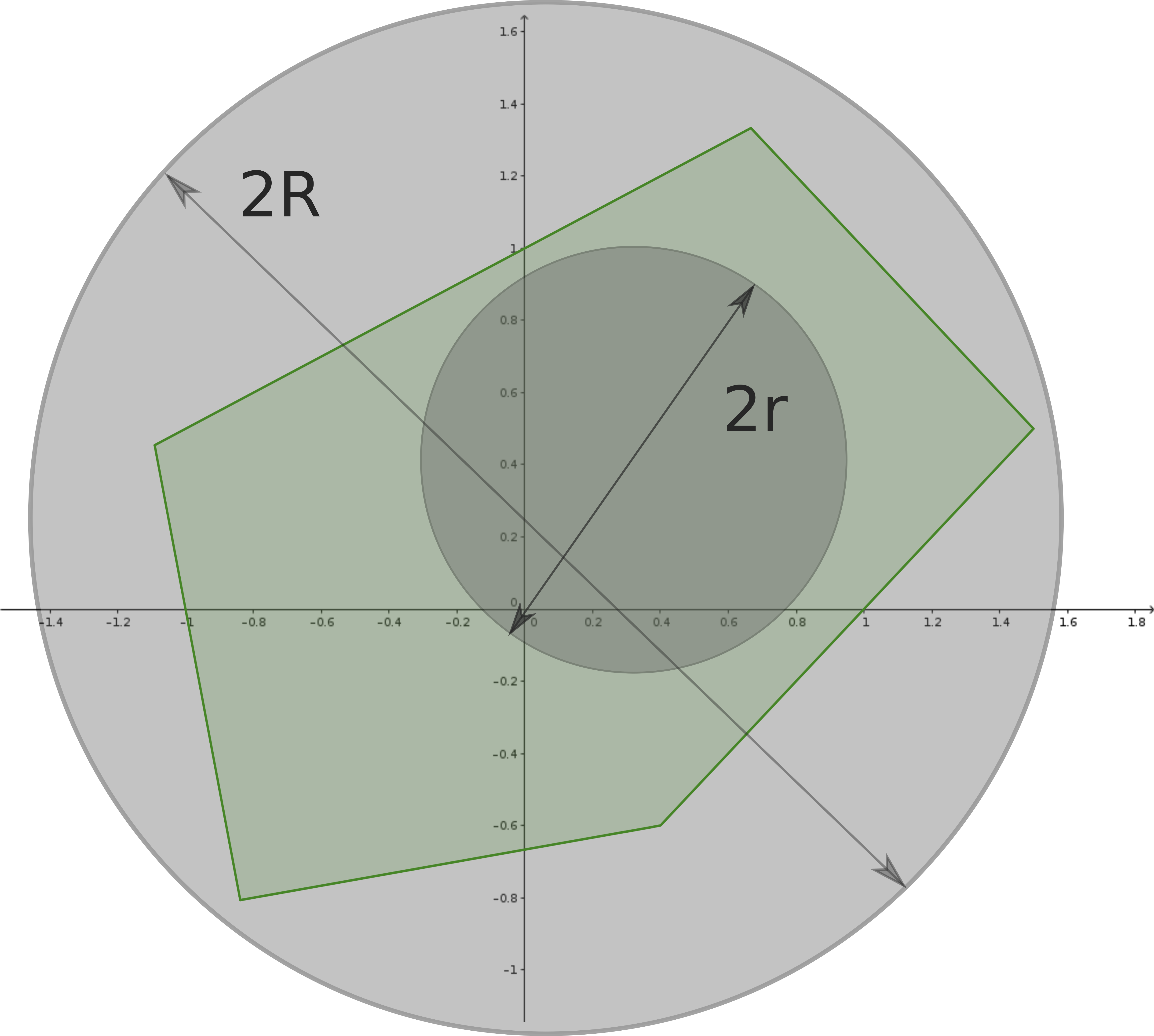}
\caption{Included and Including Balls for the feasible set of linear constraints (shown in green) \label{fig:assumption_ell}}
\end{center}
\vspace{30pt}
\end{wrapfigure} 
\vspace{-20pt}
\paragraph*{Necessary Geometric Characteristics.}
In order to know the number of steps required for the
algorithm to return an $\epsilon$-optimal solution, three scalars and a point
$x_c \in \mathbb{R}^{n}$ are needed:
\begin{itemize}
\item a radius $R$ such that 
\vspace{-15pt}
\begin{equation}
\label{eq:hypothesis_R}
X_f \subset B_{R}(x_{c}) 
\end{equation} 
\vspace{-10pt}
\item a scalar $r$ such that there exists a point $\bar{x}$ such that
\vspace{-10pt}
\begin{equation}
\label{eq:hypothesis_r}
 B_{r}(\bar{x}) \subset X_f
\end{equation} 
\item and another scalar $V$ such that
\vspace{-10pt}
\begin{equation}
\label{eq:hypothesis_V}
\underset{x \in X_f}\max ~~ f_{o} - \underset{x \in X_f}\min ~~ f_{o} \leq V.
\end{equation} 
\end{itemize}
Figure~\ref{fig:assumption_ell} illustrates the scalars $R$ and $r$. The
Feasible set (assumed to be bounded) is shown in green.
\noindent The main result can be stated as: \\
\vspace{-10pt}
\begin{myth} Assuming that $X$ is bounded, non-empty and
that scalars $R, r$ and 
\label{th:main_ellmethod}
$V$ satisfying Eqs.~\eqref{eq:hypothesis_R},~\eqref{eq:hypothesis_r} and
~\eqref{eq:hypothesis_V} are known. Then, for 
all $\epsilon \in \mathbb{R}^{*}_{+}$, the algorithm, using $N$ iterations, 
will return $\hat{x}$, satisfying: 
\vspace{-5pt}
\[ \vspace{-5pt} f_{o}(\hat{x}) \leq  f_{o}(x^{*}) + \epsilon \text{ and } \hat{x} \in X \]
With $N = 2  n  (n+1)\log \big( \frac{R}{r} \frac{V}{\epsilon} \big)$, $n$ being 
the dimension of the optimization problem.
\end{myth}
This result, when applied to LP, is the first proof of the polynomial solvability of
linear programs. This proof can be found 
in~\cite{Nemirovski2012,khachiyan1980polynomial}. \\
\begin{wrapfigure}{r}{0.54\textwidth} 
\vspace{-30pt}
  \begin{center}
\begin{cacslListing}[hbox,enhanced,drop shadow]{ACSL}
/*@ axiomatic LinAlg { 
  type vector; 
  type matrix; 
  logic vector vec_of_16_scalar(double * x) 
    reads x[0..15]; 
  logic vector vec_of_36_scalar(double * x)
    reads x[0..35]; 
  ...
  logic vector vector_add(vector A, vector B); 
  axiom vector_add_length:
    \forall vector x, y;
      vector_length(x) == vector_length(y) ==>
      vector_length(vector_add(x,y)) ==
        vector_length(x);
  axiom vector_add_select:
    \forall vector x, y, integer i;
      vector_length(x) == vector_length(y) ==>
      0 <= i < vector_length(x) ==>
      vector_select(vector_add(x,y),i) == 
      vector_select(x,i)+vector_select(y,i);
  ...
}
*/
\end{cacslListing}
\caption{ACSL Linear Algebra Theory}
\label{acsl:acsl_linalg}
  \end{center}
  \vspace{-0pt}
  \vspace{1pt}
\end{wrapfigure} 
\vspace{-35pt}
\subsection{ACSL Theory}
\noindent This section would be to describe all the manually written ACSL artifacts
to support the automatic formalization and then verification of the HLR. 
Indeed, the software analyzer takes as an input the annotated C code plus
ACSL theories that define new abstract types, functions, as well as axioms, 
lemmas and theorems. The lemmas and theorems need to be proven 
but the axioms are taken to be true. 
For the SMT solver, properties within C code as annotations are usually harder to 
prove than lemmas within an ACSL theory. On top on checking the mathematical
correctness of an ACSL annotation, the SMT solver needs to check the soundness of 
the code itself (e.g., memory allocation, function calls, ACSL contracts on the function called, etc).
Thus, the approach chosen here is to develop as much as possible the ACSL theories
to express and prove the main results used by the algorithm. 
That way, the Hoare triples at code level are only an instantiation 
of those lemmas and are relatively simple to prove.
\vspace{-10pt}
\paragraph{Linear Algebra Based ACSL Theory.} In this ACSL theory, new abstract 
types for vectors and matrices are defined. Functions allowing the instantiation
of those types of objects are also defined. As well, all of the well-known 
operations have also been axiomatized, such as vector-scalar multiplication, scalar product, norm etc. 
This ACSL theory is automatically generated during the autocoding 
process of the project and all the sizes of the vectors and matrices are known. 
Thus, within this theory, only functions that will create 
objects, from a C code pointer, of appropriate sizes (as illustrated in 
Fig.~\ref{acsl:acsl_linalg}) are being defined. ACSL code is printed in green and its keywords 
in red. The C code keywords are printed in blue and the actual C code is 
printed in black. Using the keyword \say{type}, two new abstract types, matrix and vector
are being defined. Furthermore, the ACSL constructors for those 
types and the axiomatization of vector addition is shown in Fig.~\ref{acsl:acsl_linalg}.
The first axiom states that the length of the addition of two vectors, of same length is
equal to this same length. The second axiom states that the elements of the
addition of two vectors is the addition of the elements of the two separate vectors.
Figure~\ref{acsl:acsl_linalg} shows a partial sample of the autocoded ACSL linear 
algebra theory.\\
\vspace{-20pt}	
\paragraph{Optimization and Ellipsoid Method Based ACSL Theory.} 
The axiomatization of optimization problems has already been briefly discussed in 
Section~\ref{sec:semOpt}. Additionally, work has been dedicated to axiomatize the 
calculation of the vector constraint, feasibility, epsilon-optimality, etc. 
As well, 

\begin{wrapfigure}{r}{0.52\textwidth} 
\vspace{-35pt}
  \begin{center}
\begin{cacslListing}[hbox,enhanced,drop shadow]{ACSL}
#include "axiom_linalg.h"
/*@ axiomatic Ellipsoid {
  type ellipsoid; 
  logic ellipsoid Ell(matrix P, vector x); 
  logic boolean inEllipsoid(ellipsoid E, 
  		vector z);
  ...
}
*/
\end{cacslListing}
\caption{Ellipsoid Type Definition}
\label{acsl:ellipsoid_def}
  \end{center}
  \vspace{-10pt}
\end{wrapfigure}
\noindent Ellipsoids and related properties are formally defined, 
such as those presented in Fig.~\ref{acsl:ellipsoid_def}. 
Figure~\ref{acsl:ellipsoid_def} presents the 
definition of another theory called \say{Ellipsoid} and within it, it shows
the creation of a new abstract type \say{ellipsoid} and the definition 
of the functions \say{Ell} and \say{inEllipsoid}. The function \say{Ell} returns
the Ellipsoid formed by the matrix $P$ and vector $x$ as defined in Eq.~\eqref{eq:ellipsoid_def}.
The function \say{inEllipsoid} returns true if the vector $z$ is in the Ellipsoid $E$ and returns 
false otherwise. As it was explained earlier, Theorem~\ref{th:main_ellmethod} was translated 
to ACSL and auto generated as ACSL lemmas.
All the autocoded lemmas are proven using the software analyzer Frama-C and the SMT
solver Alt-Ergo.
One of the autocoded lemmas can be found in Fig.~\ref{acsl:ellMeth_acsl_lemma}
and can be expressed using common mathematical notations as: 
\vspace{-5pt}
\begin{equation}	
\nonumber
 \color{blue} \text{Assuming:}  \color{black} ~~ 0 < \epsilon / V  < 1  
 ~~\text{;}~~ 0 < \epsilon   ~~\text{;}~~
 \forall x,y \in X,~ f_o(x) - f_o(y) \leq V   ~~\text{;}~~
\end{equation}
\vspace{-15pt}
\begin{equation}
\nonumber
\forall z \in \mathbb{R}^{n}, ~ z  \not\in  \text{Ell}(P,x) , \implies x_{best} ~ \text{is better than} ~ z 
~~\text{and}~~ \text{Vol}(\text{Ell}(P,x)) < X_{\epsilon/V}  ~~~~
 \color{blue} \text{Then:} 
 \color{black} ~~ x_{best} ~ \text{is}~\epsilon-\text{optimal}.
\end{equation}
A formal definition is also given for a point $x$ being better 
than another point $y$. The set $X_{\epsilon/V}$ represents the set obtained 
by shrinking the feasible set $X_f$ by a factor $\epsilon/V$ centered on 
the optimal point $x^{*}$. Further details about this set can be found
in~\cite{Nemirovski2012}.

\begin{figure}[ht!]
\centering
\begin{cacslListing}[hbox,enhanced,drop shadow]{ACSL Lemma}
/*@  
 lemma epsilon_solution_lemma:
	 \forall optim OPT, real r,V,epsilon, matrix P, vector x, x_best;
	  	(0 < epsilon/V < 1) ==> 0 < r ==> 0 < V ==> 0 < epsilon ==> 
		size_n(OPT) > 0 ==>
		( \forall vector x1, x2; isFeasible(OPT, x1) ==>  isFeasible(OPT, x2) ==> 
	    cost(OPT,x1) - cost(OPT,x2) <= V  ) ==>
	  ( \forall vector z; !inEllipsoid(Ell(P,x), z) ==> isBetter(OPT, z, x_best) ) ==>
	  ( \exists vector x; include(tomyset(Ell(mat_mult_scalar(ident(size_n(OPT)),r), x)) , 
					  feasible_set(OPT)) ) ==>
		 volume(tomyset(Ell(P,x))) < pow(epsilon/V*r, size_n(OPT)) ==>
		 isEpsilonSolution(OPT, x_best, epsilon);  
*/
\end{cacslListing}
\caption{Ellipsoid Method Main Lemma}
\label{acsl:ellMeth_acsl_lemma}
\end{figure}
\vspace{-10pt}
\subsection{Annotating C Code}
\noindent Details are now given about how the C code is annotated and the type
of Hoare triples used. For this, a specific 
technique was adopted. Every C code function is implemented in a separated file.
That way, for every function, a corresponding C code body (.c) file and 
header file (.h) are automatically generated.
The body file contains the implementation of the

\begin{wrapfigure}{r}{0.55\textwidth} 
\vspace{-15pt}
\begin{center}
\begin{cacslListing}[hbox,enhanced,drop shadow]{C Code + ACSL}
#ifndef getNorm_2_lib 
#define getNorm_2_lib 
#include "axiom_linalg.h"
#include "my_sqrt.h"
#include "scalarProduct_2.h"
/*@
  @ requires \valid(Ain+(0..1));
  @ ensures \result == 
  		twoNorm(vec_of_2_scalar(Ain));
  @ ensures \result >= 0;
  @ assigns \nothing;
  @ behavior Ain_non_null:
    @ assumes nonnull(vec_of_2_scalar(Ain));
    @ ensures \result > 0;
  @ behavior Ain_null:
    @ assumes !nonnull(vec_of_2_scalar(Ain));
    @ ensures \result == 0;
  @ complete behaviors Ain_non_null, Ain_null;
  @ disjoint behaviors Ain_non_null, Ain_null;
*/ 
double getNorm_2(double *Ain);
#endif
\end{cacslListing}
\caption{getNorm\_2 Header C Code File}
\label{fig:getNorm_2_header}
  \end{center}
  
    \begin{center}
\begin{cacslListing}[hbox,enhanced,drop shadow]{C Code + ACSL}
#include "getNorm_2.h"
double getNorm_2(double *Ain) {
	double sum;
	sum = scalarProduct_2(Ain, Ain);
	return my_sqrt(sum); }
\end{cacslListing}
\caption{getNorm\_2 Body C Code File}
\label{fig:getNorm_2_body}
  \end{center}
  \vspace{-30pt}
\end{wrapfigure} 
\noindent function along with annotations and loop invariants. 
The header file contains the declaration of the function with 
its ACSL contract. The first kind of Hoare triples and function 
contracts added to the code were to check basic mathematical 
operations. Figure~\ref{fig:getNorm_2_header} shows an ACSL 
contract relative to the C code function computing the 2-norm 
of a vector of a size two. The ACSL contract specifies that the 
variable returned is positive and equal to the 
2-norm of the vector of size two described by the input pointer.
Using the keywords \say{behavior}, \say{disjoint} and \say{complete} one can specify
the different scenarios possible and treat them separately.
Furthermore, it is proved that 
the result is always positive or null, and assuming the
corresponding input vector not equal to zero, the output is 
strictly greater than zero. This last property becomes important 
when one must prove there are no divisions by zero (normalizing vectors). 
\noindent The implementation of the function is presented in
Fig.~\ref{fig:getNorm_2_body}. 
Once all the C functions implementing elementary mathematical
operations have been annotated and proven, the next step
consists of annotating the higher level C functions 
such as constraint and gradient calculations, matrix and vector updates, etc.
Please find in Fig.~\ref{fig:getp_body} as an example,
the annotated C function for the function \say{getp} that computes 
the vector $p$ as described in Eq.~\eqref{eq:updateHyperplane}, needed to perform 
the ellipsoid update.
\begin{figure}[ht!]
\centering
\begin{cacslListing}[hbox,enhanced,drop shadow]{C Code + ACSL}
#include "getp.h"
void getp() {
	double norm;
	double norm_inv;
	getTranspose();
	/*@ assert mat_of_2x2_scalar(&temp_matrix[0])==
		transpose(mat_of_2x2_scalar(&P_minus[0])); */
	changeAxis();
	/*@ assert vec_of_2_scalar(&temp2[0]) ==  mat_mult_vector( 					
		mat_of_2x2_scalar(&temp_matrix[0]), vec_of_2_scalar(&grad[0])); */
	/*@ assert vec_of_2_scalar(&temp2[0]) == mat_mult_vector(transpose( 	
		mat_of_2x2_scalar(&P_minus[0]) ), vec_of_2_scalar(&grad[0])); */
	norm = getNorm_2(temp2);
	/*@ assert 1/norm == 1/twoNorm( mat_mult_vector(
		transpose( mat_of_2x2_scalar(&P_minus[0]) ), vec_of_2_scalar(&grad[0]))); */
	/*@ assert vec_of_2_scalar(&temp2[0]) == mat_mult_vector(transpose(
		mat_of_2x2_scalar(&P_minus[0])),vec_of_2_scalar(&grad[0]));  */
	norm_inv = 1.0 / (norm);
	scaleAxis(norm_inv);
	/*@ assert 1/norm == 1/twoNorm( mat_mult_vector( transpose( 
		mat_of_2x2_scalar(&P_minus[0])),vec_of_2_scalar(&grad[0]))); */
	/*@ assert vec_of_2_scalar(&temp2[0]) == mat_mult_vector(transpose( 		
		mat_of_2x2_scalar(&P_minus[0])), vec_of_2_scalar(&grad[0])); */
	/*@ assert vec_of_2_scalar(&p[0])==vec_mult_scalar(vec_of_2_scalar(&temp2[0]),1/norm); */
}
\end{cacslListing}
\caption{getp.c Body C Code File}
\label{fig:getp_body}
\end{figure}
\noindent In Fig.~\ref{fig:getp_body} one can note the presence of several function calls 
followed by ACSL annotations, encoding the corresponding specifications.
The first function call refers to the function \say{getTranspose} which computes the transpose of the 
matrix \say{P\_minus} and stores it into the variable \say{temp\_matrix}.
The function \say{changeAxis} multiplies the matrix \say{temp\_matrix} by the
vector \say{grad} and stores the result into the vector \say{temp2}. The
current state of the memory is specified at each line of code using ACSL
annotations. Then, after computing the norm and scaling the vector \say{temp2},
the annotations specify that the resulting vector stored in the variable $p$ has indeed been calculated
as stated in Eq.~\eqref{eq:updateHyperplane}.

%% file: online.tex
\section{Sequential Optimization Problems}
\label{sec:online}
\noindent In this section, convergence guarantees are provided
for a class of optimization problems used online. 
For this, an optimization problem with parameterized constraints 
and cost will be considered.
This section concerns the study of how this 
parameter affects the optimization problem at each iteration and 
how to find the condition that the parameter needs to satisfy in order to prove
convergence for every point along the trajectory. \\
First, the study focuses on the special case of linear constraints. Following this, 
another section will be dedicated to SOCP constraints.
\subsection{Parameterized Linear Constraints}
\label{sec:param_lin_constr}
\noindent 
The objective is to solve in real-time the optimization problem described 
in Eq.~\eqref{opt:original}. The vector $X \in \mathbb{R}^{n_x} $ denotes the 
decision vector.
\begin{equation}
\setlength{\jot}{-2pt}
\begin{aligned}
&  \underset{X}{\text{minimize}}
& & 	 f_{o}(X) \\
& \text{subject to} & & AX \leq b \\
\end{aligned}
\label{opt:original}
\end{equation}
The initialization of this optimization problem is 
done using Eq.~\eqref{eq:initEquality}, where $S$ is a full rank 
matrix. Usually, $S$ represents a selector matrix and is of the form: 
$[I_p ~ O_{p \times ( n_x - p ) }]$. A selector matrix is a matrix
that selects one or more component from $X$. If $S$ is a selector matrix
then $SX$ returns certain components of $X$.
In that case it is obviously full
rank. $x_o$ denotes the input of the controller and it is written $\hat{x}_{o}$ 
to account for the fact that $x_{o}$ changes from one optimization 
problem to another.
\vspace{-5pt}
\begin{equation}
S X = \hat{x}_{o}
\label{eq:initEquality}
\vspace{-5pt}
\end{equation}
One can decompose and separate the equality and inequality constraints hidden 
behind the matrix $A$ and vector $b$. That way, Eq.~\eqref{opt:original}
can be written as:	
\begin{equation}
\setlength{\jot}{-1pt}
\label{opt:decomposed}
\begin{aligned}
&  \underset{X}{\text{minimize}}
& & f_{o}(X) \\
& \text{subject to} & & A_{\rm eq}  X = b_{\rm eq} \\
& & & A_{\rm ineq} X \leq b_{\rm ineq} \\
& & & S  X = \hat{x}_{o} 
\end{aligned}
\end{equation}
The idea here is to project all the equality constraints in order to 
eliminate them while keeping track of the variable parameter, $\hat{x}_{o}$. 
There exist matrices $M$, $A_1$ and $A_2$ such that for all
vectors $X$ satisfying the equality constraints of the problem defined by Eq.~\eqref{opt:decomposed}, 
there exists a vector $Z$ such that:
\vspace{-5pt}
\begin{equation}
\label{eq:lin_proj}
X = A_1  b_{\rm eq} + A_2  \hat{x}_{o} +  M  Z
\vspace{-5pt}
\end{equation}
Equation~\eqref{eq:lin_proj} is a direct implication that the set of solutions of a linear 
system is an affine set. 
In this same equation, $M$ is a matrix formed by an orthonormal basis of the null space of the matrix
$ \begin{bmatrix} A_{\rm eq} \\[-5pt]  S
\end{bmatrix}$ and $d$ denotes the total number of equality constraints
(number of rows of $A_{\rm eq}$ + number of rows of $S$). 
The vector $Z$ then belongs to $\mathbb{R}^{n_z}$ with $n_z = n_x - d$. Thus, the original 
optimization problem described by Eq.~\eqref{opt:original} is equivalent to the projected problem:
\begin{equation}
\setlength{\jot}{-1pt}
\label{opt_projected}
\begin{aligned}
&  \underset{Z}{\text{minimize}}
& & f_{o}(A_1  b_{\rm eq} + A_2  \hat{x}_{o} +  M  Z) \\
& \text{subject to} & & A_{\rm f} Z \leq b_{o} + Q  \hat{x}_{o}  \\
\end{aligned}
\end{equation}
\vspace{-10pt}
With:
\begin{equation}
A_{\rm f} = A_{\rm ineq}  M
~~~ \text{;} ~~~~~
b_{o} =  b_{\rm ineq} -A_{\rm ineq}  A_{1}  b_{\rm eq}
~~~~~~ \text{and} ~~~~~~~~
Q = -A_{\rm ineq} A_{2}.
\end{equation}
More details and references about equality constraints elimination can be found 
in~\cite{boyd2004convex}. Using the Ellipsoid Method online to solve this optimization problem
requires the computation of geometric characteristics on the feasible set of this parametric 
optimization problem for every possible $x_{o}$. For that reason, for now, let us assume: 
$\norm{x_{o}}_{2} ~\leq~ r_{o}$.\\
\noindent The parameterized polyhedral set below $P_{ \hat{x}_{o} }$ is defined as:
\vspace{-5pt}
\begin{equation}
P_{ \hat{x}_{o} } = \{ z \in \mathbb{R}^{n_z} : 
A_{\rm f}  z \leq  b_{o} + Q \hat{x}_{o} \}
\vspace{-5pt}
\end{equation}
Having this collection of polyhedral sets, the goal is to compute 
ball radii that will tell us about the volume of 
the feasible set that would be true for every $x_{o}$.
The operator $\nu(.)$ returns for a matrix $A$ the vector $\nu(A)$ whose coordinates 
are the 2-norm of the rows of $A$. 
Thus, a way of computing those geometric characteristics is to consider the two extreme polyhedral 
sets below:
\begin{equation}
P_{\rm min} = \{ z \in \mathbb{R}^{n_z} : A_{\rm f}  z \leq  b_{o} - r_{o}  \nu(Q) \},
\end{equation}
\begin{equation}
P_{\rm max} = \{ z \in \mathbb{R}^{n_z} : A_{\rm f}  z \leq  b_{o} + r_{o}  \nu(Q) \}.
\end{equation}
\begin{wrapfigure}{r}{0.50\textwidth} 
\vspace{-30pt}
\begin{center}
\includegraphics[scale=0.45]{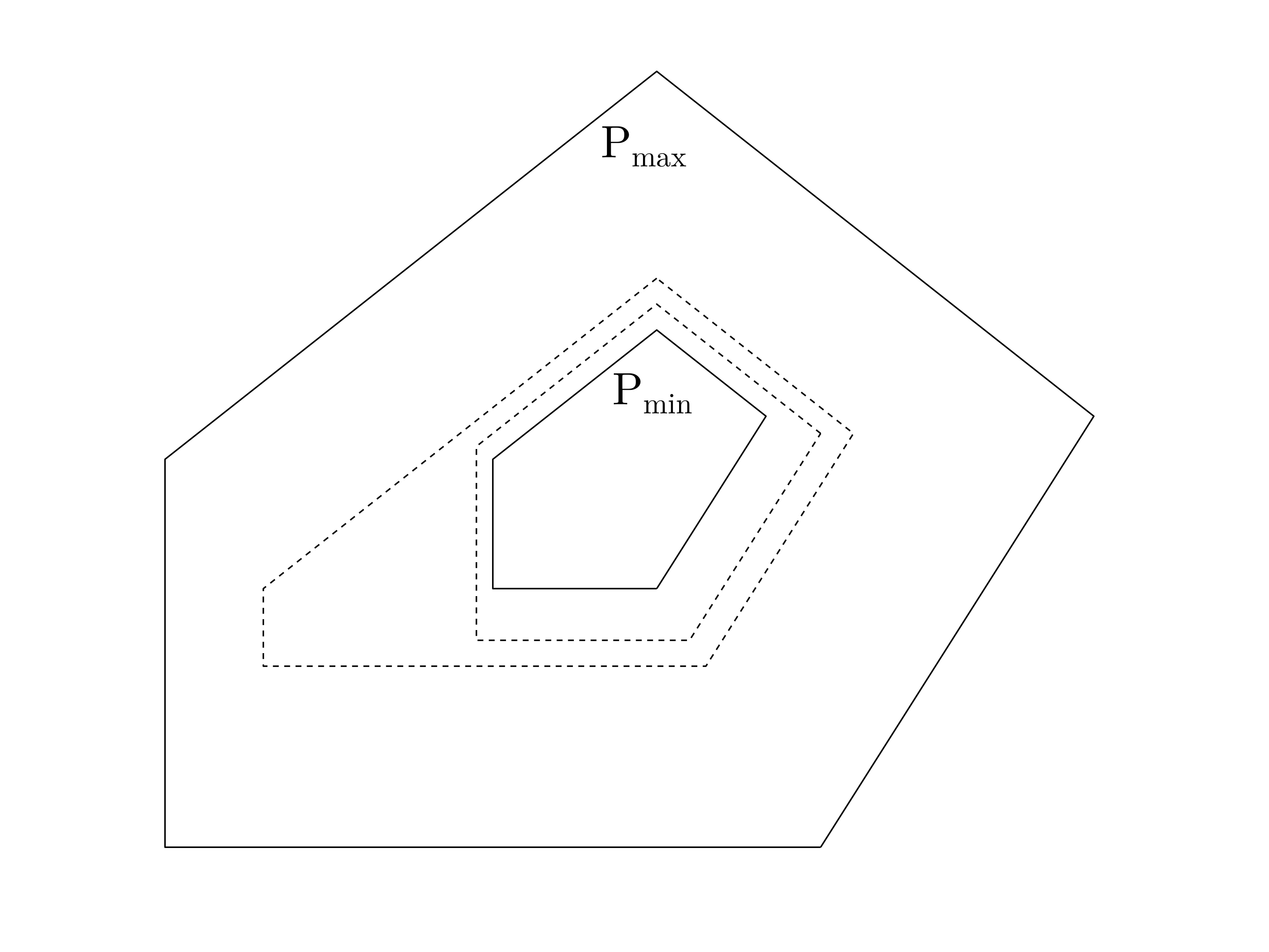}
\caption{$P_{\rm max}$ and $P_{\rm min}$ Polyhedral Sets}
\label{fig:poly_u_l}
\end{center}
\vspace{-10pt}
\end{wrapfigure} 
In order to give an example of this concept, please find in Fig.~\ref{fig:poly_u_l} 
an illustration of such polyhedral sets (the illustrated sets have no 
physical meaning and do not represent any MPC problem).
The two extreme polyhedral sets are drawn with solid lines and
the actual feasible polyhedral sets are drawn using dotted lines.
The values used are: \\
\vspace{10pt}
\[ A_{\rm f} =  {\setlength\extrarowheight{-10pt}    
\begin{bmatrix}
-1  &  1 \\
1  &  1 \\
1  & -0.5 \\
0  &  1 \\
-1 &  0 \\
0  & -1
\end{bmatrix}}
~~~, \hspace{2pt}
\nu(Q) = {\setlength\extrarowheight{-10pt}
\begin{bmatrix}
 1 \\
 1 \\
 1 \\
 1 \\
 1 \\
 1 \\
\end{bmatrix} } ~~~, \hspace{2pt}
 b_{o} = {\setlength\extrarowheight{-10pt}
\begin{bmatrix}
 1 \\
 2 \\
 1 \\
 1.5 \\
 0.5 \\
 0.5 \\
\end{bmatrix} }  ~~~ \text{and} ~~~
r_o = 0.5.   \]
\begin{myfact}{ [Extreme Polyhedral Sets] }
\vspace{-5pt}
\begin{equation}
\nonumber
\forall x_{o} \in \mathbb{R}^{n}~\text{s.t.}~ \norm{x_{o}}_{2} ~\leq~ r_{o}~~,~~ P_{\rm min} \subset P_{\hat{x}_{o}} \subset P_{\rm max}.
\end{equation}
\end{myfact}
\begin{po}
Take $x_{o}$ such that $\norm{x_{o}}_{2} ~ \leq ~ r_{o}$. \\
First, using the Cauchy-Schwarz inequality, one can write:
\vspace{-5pt}
\begin{equation}
\abs{ (Q   x_{o}) (i)}  ~=~  \abs{ \text{row}(Q,i)^{T}   x_{o} }  ~\leq~ 
\norm{\text{row}(Q,i)}_{2}  \norm{x_{o}}_{2}  ~\leq~ \nu(Q)(i) r_{o} ~~~ \forall i
\vspace{-5pt}
\end{equation}
\begin{equation}
\vspace{-10pt}
\nonumber
\Longrightarrow -\nu(Q) r_{o} \leq  Q x_{o}  \leq  \nu(Q)  r_{o}.
\vspace{5pt}
\end{equation}
Then, if $x \in P_{\rm min}$, one can conclude that 
$ A_{\rm f}  x \leq  b_{o} - r_{o}  \nu(Q) $.
Using the inequality above, it is clear that it implies $x \in P_{x_{o}}$. Similarly, 
assuming that $x \in P_{x_{o}}$ and using the inequality below, it is clear that $x \in P_{\rm max}$.
\QEDB
\end{po}
\noindent Next, one needs to find three scalars $r, R$ and $V$ such that:
\begin{equation}
\exists ~ \bar{z_{1}} ~~\text{such that}~~ B(\bar{z_{1}}, r) \subset P_{\rm min} ~~,
\label{eq:r_cond_mpc}
\vspace{-7pt}
\end{equation}
\begin{equation}
\exists ~ \bar{z_{2}} ~~\text{such that}~~ P_{\rm max} \subset B(\bar{z_{2}}, R) ~~,
\label{eq:R_cond_mpc}
\vspace{-5pt}
\end{equation}
\begin{equation}
V \geq  \underset{z \in P_{\hat{x}_{o}}}{\text{max}} ~~ f_{o}(z) - \underset{z \in P_{\hat{x}_{o}}}{\text{min}} ~~ f_{o}(z) ~~,~ \forall  ~~ \norm{\hat{x}_{o}} \leq r_{o}.
\label{eq:V_cond_mpc}
\end{equation}
For the first scalar, $r$, one can compute a numerical value by running an 
off-line optimization problem finding the largest ball inside $P_{\rm min}$. If no solution
can be found, the value of $r_o$ needs to be decreased, and the process is repeated until 
acceptable values for $r_o$ and $r$ are found. Further information 
about finding the largest ball in a polytope can be found in~\cite{boyd2004convex}. \\
As a consequence of equality constraint elimination, and seen in Eq.~\eqref{eq:lin_proj}, 
the relation between the original decision vector $X$ and the
projected vector $Z$ can be written as:
\begin{equation}
X = A_{\rm proj}  \begin{bmatrix} 
    b_{\rm eq} \\ \hat{x}_{o} 
\end{bmatrix}  + M Z.
\label{eq:proj_total}
\end{equation}
The decision vector $X$ can now be decomposed into two parts, $\textbf{x}$ and $\textbf{u}$. The
part $\textbf{u}$ is bounded due to constraints in the original optimization problem (described by
Eq.~\eqref{opt:original}).
If no constraints on \textbf{u} were originally present, one can construct bounds on the 
projected vector $Z$ to make the problem bounded and simpler to analyze. 
The point here being that from bounded variables within the 
vector $X$, one can infer bounds on the projected vector $Z$. There is no need to have 
original bounds specifically on the collection of future inputs. Rewriting 
Eq.~\eqref{eq:proj_total} yields:
\begin{equation}
\begin{bmatrix}
    \textbf{x} \\ \textbf{u} 
\end{bmatrix} = \begin{bmatrix}
    A_{11} & A_{12} \\
    A_{21} & A_{22}
\end{bmatrix} 
\begin{bmatrix}
    b_{\rm eq} \\
    \hat{x}_{o}
\end{bmatrix}
 + 
 \begin{bmatrix}
    M_{1} \\
    M_{2} 
\end{bmatrix}
Z.
\end{equation}
Following this, one can conclude:
\begin{equation}
 Z = M_{2}^{-1}  \big( \textbf{u} - A_{21} b_{\rm eq} - A_{22} \hat{x}_{o}  \big).
\end{equation}
Therefore assuming again that $\norm{\hat{x}_{o}} \leq r_{o}$, one can compute a value of $R$ such that:
\begin{equation}
\norm{Z} \leq  \norm{M_{2}^{-1}}   \big( \norm{\textbf{u}} + \norm{A_{21} b_{\rm eq}} + \norm{A_{22}} r_o  \big) = R.
\label{eq:R_mpc}
\end{equation}
On the other hand, from the physical meaning of the variables and the constraints of the 
optimization problem, one can construct bounds in which the variables should live, and 
therefore find a lower bound for $V$.  \\
With the values $R$ and $r$, one can now guarantee the convergence of a family of optimization problems parameterized by $x_o$.   
\begin{myth}{ [MPC Ellipsoid Method Convergence] \\}
The problem given by Eq.~\eqref{opt:decomposed} is run in an online
fashion in order to implement receding horizon control. \\
Using the Ellipsoid Method and initializing the first Ellipsoid by $B(\bar{z_{2}}, R)$, 
the method will find an $\epsilon$-solution using $N$ iterations for all $\hat{x}_{o}$ such that 
$\norm{\hat{x}_{o}} \leq r_{o}$, with:
\vspace{-5pt}
\[ N = 2 n(n+1)  
\log \Big(  \frac{R}{r} \frac{V}{\epsilon} \Big) \vspace{-5pt} \]
with the variables $\bar{z_{2}}$, $r$, $R$ and $V$ satisfying the Eqs.~\eqref{eq:r_cond_mpc},
~\eqref{eq:R_cond_mpc} and~\eqref{eq:V_cond_mpc}. The variable $n$ denotes 
the dimension of the parameterized optimization problem.
\end{myth}
\begin{po}{.}
Let us assume, the problem given by Eq.~\eqref{opt:decomposed} is run online.
For each iteration, the input parameter $x_{o}$ is assumed to satisfy $\norm{x_{o}} \leq r_{o}$. 
$X_{\rm f}$ denotes the feasible set of the current optimization
problem to solve. That way, Eq.~\eqref{eq:r_cond_mpc} is satisfied. Thus, by definition 
of $r$, we know that:
\vspace{-5pt}
\begin{equation}
\nonumber
\exists \bar{z_1} ~~ \text{such that}~~ B(\bar{z_1}, r) \subset  P_{\rm min}  \subset X_f .
\vspace{-5pt}
\end{equation}
Similarly, because it is assumed that $R$ satisfies Eq.~\eqref{eq:R_cond_mpc}, the 
following property holds:
\vspace{-5pt}
\begin{equation}
\nonumber
\exists \bar{z_2} ~~ \text{such that}~~ X_f \subset B(\bar{z_1}, r).
\vspace{-5pt}
\end{equation}
As well, the scalar $V$ satisfies Eq.~\eqref{eq:hypothesis_V}.
Finally, One can conclude that the returned point will indeed 
be $\epsilon$-optimal using $N$ iterations thanks to 
Theorem~\ref{th:main_ellmethod}.
\QEDB
\end{po}
\noindent \emph{Linear Programs:} In the case of Linear Programs, the method developed is identical and
having a linear cost, the scalar $V$ is easily found by running the two optimization 
problems below.
\begin{equation}
\nonumber
\setlength{\jot}{-2pt}
\label{find_V}
\begin{aligned}
&  \underset{\hat{x}_{o}, z}{\text{minimize}}
& & c^{T}  (A_{1}b_{\rm eq} + A_{2}\hat{x}_{o}  + Mz) \\
& & & A_{\rm f}  z \leq b_{o} + Q  \hat{x}_{o}  \\
& & & \norm{\hat{x}_{o}}_{2} ~ \leq ~ r_{o} \\
\end{aligned}
~~~~~~~~~~
\setlength{\jot}{-2pt}
\begin{aligned}
&  \underset{x_{o}, z}{\text{maximize}}
& & c^{T}  (A_{1}b_{\rm eq} + A_{2}x_{o}  + Mz) \\
& & & A_{\rm f}  z \leq b_{o} + Q  x_{o}  \\
& & & \norm{x_{o}}_{2} ~ \leq ~ r_{o} \\
\end{aligned}
\end{equation}
\vspace{-20pt}
\subsection{Second-Order Conic Constraints}
\noindent In the last section, the fact that only linear constraints was
present was used to find the radius $r$. Therefore, in this 
section we aim to give ways of computing these
constants for second-order constraints. Consider that 
one wants to solve a model predictive control optimization problem 
given by Eq.~\eqref{opt:socp_original}.
\begin{equation}
\label{opt:socp_original}
\setlength{\jot}{-2pt}
\begin{aligned}
&  \underset{X}{\text{minimize}}
& & f_{o}(X) \\
& & & \norm{A_{i}  X + b_{i} }_{2} ~ \leq ~ c_{i}^{T} X + d_{i} ~~,~i=1 \dots m \\
& & & A_{\rm eq} X = b_{\rm eq}  \\
& & & S  X = \hat{x}_{o}
\end{aligned}
\end{equation}
We assume that no equality constraints are hidden in the second-order constraints
(if not, a very simple analysis will confirm this and one can extract those
equality constraints and put it into the couple $(A_{\rm eq}, b_{\rm eq})$ 
from equation~\eqref{opt:socp_original}). Performing again an equality 
constraint elimination, we end up with an equivalent optimization problem, of smaller dimension, 
that contains no equality constraint.
%
%
We want to find the radius of the biggest ball inside the feasible set of 
this latter problem. Unfortunately, second-order constraints are still present and
finding the largest balls inside
second-order cones is not an easy task. For this, we use the equivalence of 
norms in finite dimensions, noting that $\norm{ \cdot }_{1}$ 
and $\norm{ \cdot }_{\infty}$ are linear, 
to perform a linear relaxation on the second-order constraints and end up with 
a polyhedral set as the feasible set. For instance, the problem described by
Eq.~\eqref{opt:socp_original_lin} represents a linear relaxation of the original problem 
stated in Eq.~\eqref{opt:socp_original} after having eliminated the equality constraints.
\begin{equation}
\label{opt:socp_original_lin}
\setlength{\jot}{-2pt}
\begin{aligned}
&  \underset{X}{\text{minimize}}
& & f_{o}^{'}(Z) \\
& & & \sqrt{n}  \norm{A_{i}^{'}  Z + b_{i}^{'} }_{\infty} ~ \leq
 ~ c_{i}^{'T} Z + d_{i}^{'} ~~,~i=1 \dots m \\
\end{aligned}
\end{equation}
Thus, by finding the largest ball inside the resulting feasible set (which is a polyhedral set),
a ball is finally found inside the original second-order cone.

%% file: correction.tex
\vspace{-10pt}
\section{Bounding the Condition Number}
\label{sec:correction}
\noindent  The use of the Ellipsoid Method was justified by arguing that
it represents a trade off between performance and safety. As well, it also 
attracted our attention for its numerical properties. Although at first,
the Ellipsoid Method appears to be numerically stable, finding {\em a priori}
bounds on the program's variables is challenging.
The worst case occurred when the
separating hyperplane has the same direction at each iteration. 
In this case, the condition number of the successive ellipsoids increases
exponentially, and so do the program's variables. This 
worst case is very unlikely to happen in practice. Nevertheless, a mathematical 
way to get around this case is needed, along with a way to compute {\em a priori}
bounds on the variables before the execution of the program.
In this section, a method for controlling the condition number 
of the successive ellipsoids and a way to compute those bounds by adding a 
correcting step in the original algorithm is presented.
This section also includes how the formal proof of the corresponding software is modified to support 
the correctness of the resulting modified ellipsoid algorithm.
\subsection{Bounding the Singular Values of the Successive Ellipsoids}
\label{sec:bounding}
\begin{wrapfigure}{r}{0.5\textwidth} 
\vspace{-45pt}
\includegraphics[width=0.5\textwidth]{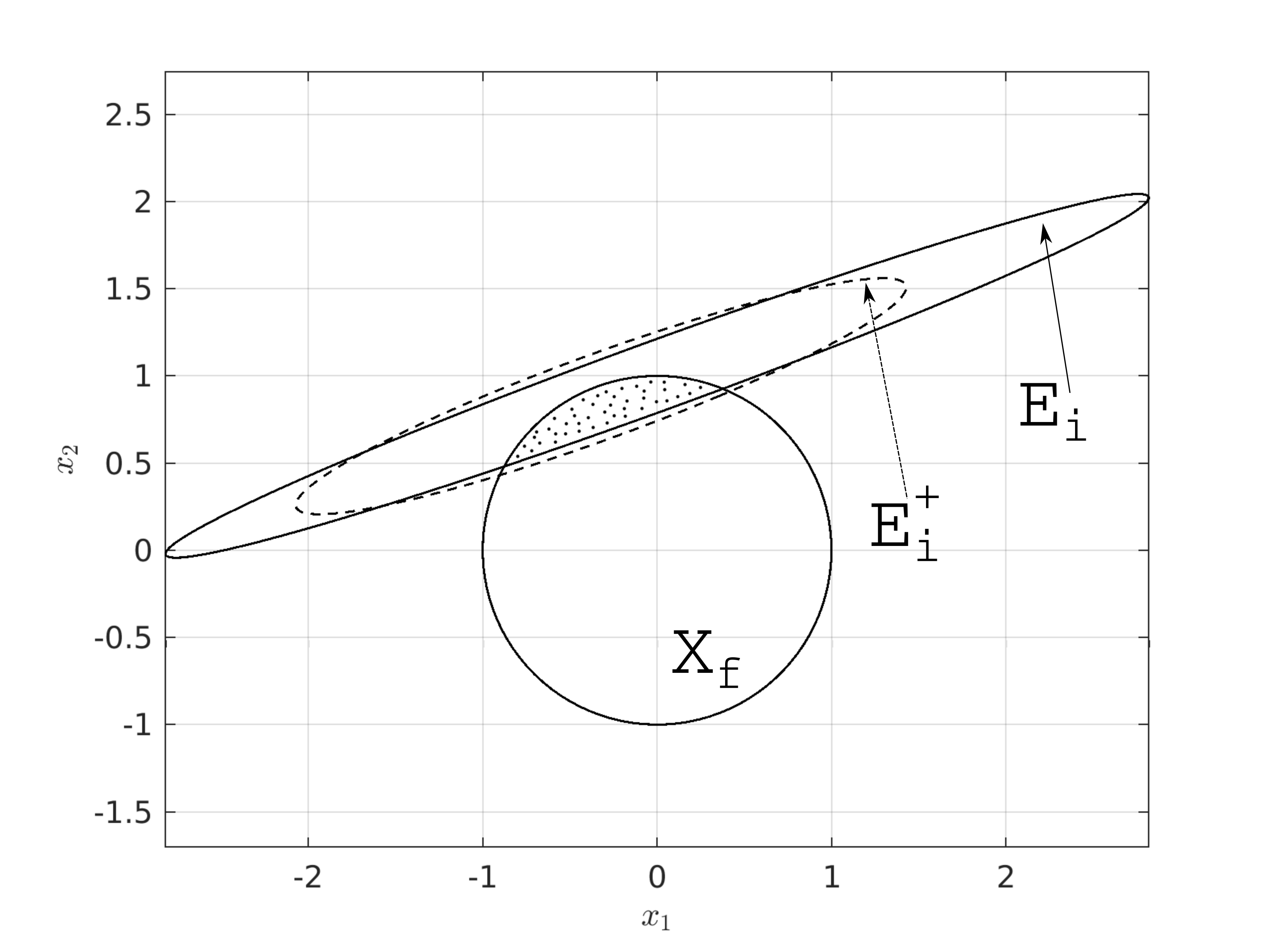}
\caption{Corrective step for Ellipsoid $E_i$ according to Eq.~\eqref{eq:update_corrected}. 
The feasible set $X_f$ is illustrated as the unit ball. \label{fig:ellWide2}}
\vspace{30pt}
\end{wrapfigure}

\noindent  When updating $B_{i}$ by the usual formulas of the ellipsoid 
algorithm (Eq.~\eqref{eq:updateShape}), $B_{i}$ evolves according to 
$B_{i+1} = B_{i} D_{i}$, where $D_{i}$ has $n-1$ singular values equal 
to $n / \sqrt{n^{2}-1}$, and has
one singular value equal to $n/(n + 1)$. 
It follows that at a single step the largest and the smallest
singular values of $B_{i}$ can change by a factor from 
$[1/2,~2]$.
The objective is to prove that one can modify the algorithm to
bound the singular values of the matrix $B_{i}$
throughout the execution of the program. \\
\emph{Minimum Half Axis:}  First, we claim that if $\sigma_{min}(B_i)$ is 
less than than $ r \epsilon/V$ then the algorithm has already found an
$\epsilon$-solution. Note that the scalar $\epsilon$ is the desired precision and the scalars $r$ and 
$V$ are defined in Section~\ref{sec:cvxEll}. Please find below a proof of this 
statement.
\begin{po}
Let us assume $\sigma_{min}(B) < r \epsilon /V$. In this case, $E_{i}$
 is contained in the strip between two parallel hyperplanes, the width of the 
strip being less than $2 \cdot r \epsilon /V$ and consequently $E_{i}$ does not contain
 $X_{\epsilon} = \theta X_f + (1-\theta) x_{*}$, where $x_{*}$ is the minimizer 
 of $f_{o}$ and $\theta = \epsilon/V$ (because $X_{\epsilon}$ contains a ball 
 of radius $ r \epsilon/V$). Consequently, there exists $z \in X_f$ such that
$y = \theta z + (1-\theta) x_{*} \in X_{\epsilon}$ but $\notin E_{i}$ , implying 
by the standard argument (developed in~\cite{Nemirovski2012}) that the best value $f^{+}$ of $f$ processed so far 
 for feasible solutions satisfies $f^{+} \leq f(y) \leq f(x_{*}) + \theta f(z-x_{*})$ 
which implies that $f^{+}  \leq f(x_{*}) + \epsilon $. We can thus stop the 
algorithm and return the current best point found (feasible and smallest cost). 
\QEDB
\end{po}
\emph{Maximum Half Axis:} We argue now that one can modify the original 
ellipsoid algorithm in order to bound the value of the maximum singular 
value of $B_i$. 
When the largest singular value of $B_i$ is less than $2 R\sqrt{n+1}$, 
we carry out a step as in the basic ellipsoid method. When this singular 
value is greater than $2 R\sqrt{n+1}$, a corrective step is applied to $B_i$, 
which transforms $E_i$ into $E_i^{+}$. 
Under this corrective step, $E_i^{+}$ is a localizer along with $E_i$, specifically 
$E_i \cap X_f \subset  E_{i}^{+} \cap X_f$, and additionally:
\renewcommand\labelitemi{}
\begin{itemize}
\item (a) The volume of $E_{i}^{+}$ is at most $\gamma$ (upper bound of reduction ratio 
from Eq.~\eqref{eq:volume_ratio}) times the volume of $E_i$ ;
\item (b) The largest singular value of $B_{i}^{+}$ is at most $2R\sqrt{n + 1}$.
\end{itemize}
The corrective step is described as follows. First, define
$\sigma = \sigma_{max}(B_{k})> 2R\sqrt{n+1} $ and let 
$e_i$ be a unit vector corresponding to the singular direction. We then consider:
\vspace{-5pt}
\[ G = \text{diag} \bigg( \sqrt{n/(n+1)}, ~ \sqrt{n+1}/\sigma, ~ \dots, ~ 
\sqrt{n+1}/\sigma \bigg). \vspace{-5pt} \]
This case is concluded by performing the update:
\vspace{-5pt}
\begin{equation}
\label{eq:update_corrected}
B_{i+1} = B_{i} \cdot G ~~~ \text{and} ~~~
c_{i+1} = c_{i} -(e_{i}^{T}c_{i})\cdot e_{i}.
\vspace{-5pt}
\end{equation} 
Please find in Fig.~\ref{fig:ellWide2} an illustration of this corrective step. 
The ellipsoid $E_i$ shown in Fig.~\ref{fig:ellWide2} appears to have a condition number
too high due to its large semi-major axis. The correction step as detailed previously has
been performed and the corrected ellipsoid $E_{i}^{+}$ is shown with dotted line.
As one can see, the volume as been decreased and the area of interest $E_i \cap X_f$
(dotted area) still lies within the ellipsoid $E_{i}^{+}$.
Hence, one can conclude that throughout the execution of the code:
\[ \sigma_{min}(B_i) \geq \frac{1}{2} \frac{r \epsilon}{V} = \frac{r \epsilon}{2V} 
~~~~ \text{and} ~~~~
 \sigma_{max}(B_i) \leq  2  \times 2R \sqrt{n+1}  = 4R \sqrt{n+1}.  \]
%
\subsection{Corresponding Condition Number}
\noindent From the definition of the condition number, one can write:
\vspace{-15pt}
\begin{equation}
\nonumber
k(B) = \norm{B} \cdot  \norm{B^{-1}} = \sigma_{max}(B) / \sigma_{min}(B).
\end{equation} 
%
Thus, by bounding the singular values of $B$, a bound on its condition number
can be constructed. The factors $1/2$ and $2$ that could affect the singular values 
of $B$ at each iteration are also taken into account to get:
\vspace{1pt}
\[ k(B) \leq  \Bigg( \frac{2}{1/2} \cdot \cfrac{ 2 R \sqrt{n+1} }{r \epsilon / V} \Bigg) = 
\Bigg( \cfrac{ 8 R \sqrt{n+1} }{r \epsilon / V} \Bigg) \vspace{-5pt} \]
and
\[ \vspace{-5pt} \norm{B}  = \sigma_{max}(B) \leq  4 R \sqrt{n+1} .  \]
\vspace{-10pt}
\subsection{Corresponding norm on $c$}
\noindent At each iteration it is known that the optimal point belongs to the
current ellipsoid. Thus:
\vspace{-5pt}
\[  \norm{ x^{*} - c_{k} } ~ = ~ \norm{  B_{k} u }  ~ \leq ~ 
\norm{B_{k}} \cdot \norm{u } ~ \leq ~ \norm{B_k}, 
 ~~~~\text{for some } u \in B_1(0). \vspace{-5pt} \]
Finally,
\vspace{-5pt}
\[  \norm{c_{k}} ~\leq~  R + \norm{x_c} + \norm{B_k}.  \vspace{-5pt} \]
\subsection{Consequences on the Code}
\noindent In this section, we explain how to implement this correcting step and give the tools to 
verify it.
In order to detect an ellipsoid with large semi-major axis, the largest 
singular value $\sigma_{max}$ of the current matrix $B_k$ needs to be computed. However, performing a singular 
value decomposition would be far too expensive and slow (this decomposition being performed 
at each iteration).  
The Frobenius norm, defined in Def.~\ref{def:frob}, which is a
well-known upper bound on the maximum singular value, is computed instead.
%
\begin{mydef}
\vspace{-5pt}
\label{def:frob}
The Frobenius norm of a matrix $A \in \mathbb{R}^{n \times n}$ is:
\[ \norm{A}_{F} ~ = ~ \sqrt{ \sum_{i=1}^{n} \sum_{j=1}^{n} a_{i,j}^{2}  } \]
\vspace{-15pt}
\end{mydef}
%
\noindent The Frobenius norm of a vector has been 
axiomatized by setting it equal to the vector 2-norm 
of the \say{vectorized} matrix. A matrix is \say{vectorized} by concatenating all its 
rows in a single vector. 
In the case of an overly large semi-major axis, the direction $e$ in which this axis lies
is needed as well. Therefore, the power iteration algorithm is performed in order 
to compute this information.

%% file: flp.tex
\section{Floating-Point Considerations}
\label{sec:flp}
\noindent Standard notation is used for rounding error 
analysis~\cite{rump2006,rump2012,roux2015}, $\fl(·)$ being the result 
of the expression within the parenthesis rounded to the nearest floating-point number. 
The relative rounding error unit is written \textbf{u} and \textbf{eta} denotes 
the underflow unit. For IEEE 754 double precision (binary64) we have \textbf{u}$=2^{-53}$ 
and \textbf{eta}$=2^{-1074}$. \\
This section presents an analysis targeting the numerical properties 
of the ellipsoid algorithm. Contributions already have been made concerning 
finite-precision calculations within the ellipsoid method~\cite{khachiyan1980polynomial}. 
However, this work only shows that it is possible to compute approximate solutions without
giving exact bounds, and is only for Linear Programming (LP). 
Also, the analysis performed considers abstract 
finite-precision numbers and the actual machine floating-point types are not mentioned. Thanks to the 
analysis performed in this section, using the IEEE standard for floating-point 
arithmetic and knowing exactly how the errors are being propagated, it is possible to check
{\em a posteriori} the correctness of the analysis using static 
analyzers~\cite{goubault2011static,putot2004static}.
\vspace{-15pt}
\subsection{Preliminaries}
\noindent Within this algorithm, we focus our attention on the update formulas 
\eqref{eq:updateCenter}, \eqref{eq:updateShape} and \eqref{eq:updateHyperplane}, 
allowing us to update the current ellipsoid into the next one. 
In order to propagate the errors due to rounding through the code,
we state now a useful theorem dealing with matrix perturbations and inverses.
\begin{myth}{\cite{Ghaoui2002}[Matrix perturbations and Inverse]}
Let $A$ be a non-singular matrix of $\mathbb{R}^{n \times n}$ and 
$\Delta A$ a small perturbation of $A$. Then,
\begin{equation}
\vspace{-15pt}
\frac{\norm{(A+\Delta A)^{-1} - A^{-1}}}{\norm{A^{-1}}} \leq 
k(A) \frac{\norm{\Delta A}}{\norm{A}}
\label{eq:inv_pert}
\end{equation}
\vspace{-5pt}
\label{th:inverse_error}
\end{myth}
\subsection{Norms and Bounds}
\noindent To successfully perform the algorithm's numerical stability analysis, we need 
to know how \say{big} the variables can grow within the execution of the 
algorithm. Indeed, for a given computer instruction, the errors due to floating-point 
arithmetic are usually proportional to the variables values. 
As it was explained in Section~\ref{sec:correction}, 
we slightly modified the ellipsoid algorithm to
keep the condition numbers of the ellipsoid iterates under control.
Therefore, implementing this corrected algorithm, we can use the following
results:
\begin{equation}
\norm{p}  =  \norm{B^{T}e} / \sqrt{e^{T}BB^{T}e}  = 1,
\vspace{-5pt}
\end{equation}
\vspace{-10pt}
\begin{equation}
\norm{B}  \leq 4R\sqrt{n+1},
\vspace{-5pt}
\end{equation}
\vspace{-10pt}
\begin{equation}
k(B) \leq (8RV\sqrt{n+1})/(r \epsilon),
\vspace{-5pt}
\end{equation}
\vspace{-10pt}
\begin{equation}
\norm{c} \leq R + \norm{x_c} + \norm{B},
\end{equation}
where $n,R,r,V, x_c$ and $\epsilon$ are the variables described in Section
\ref{sec:cvxEll}.
\vspace{-10pt}
\subsection{Problem Formulation and Results}
\begin{wrapfigure}{r}{0.5\textwidth} 
\vspace{-50pt}
\begin{center}
\includegraphics[width=0.4\textwidth]{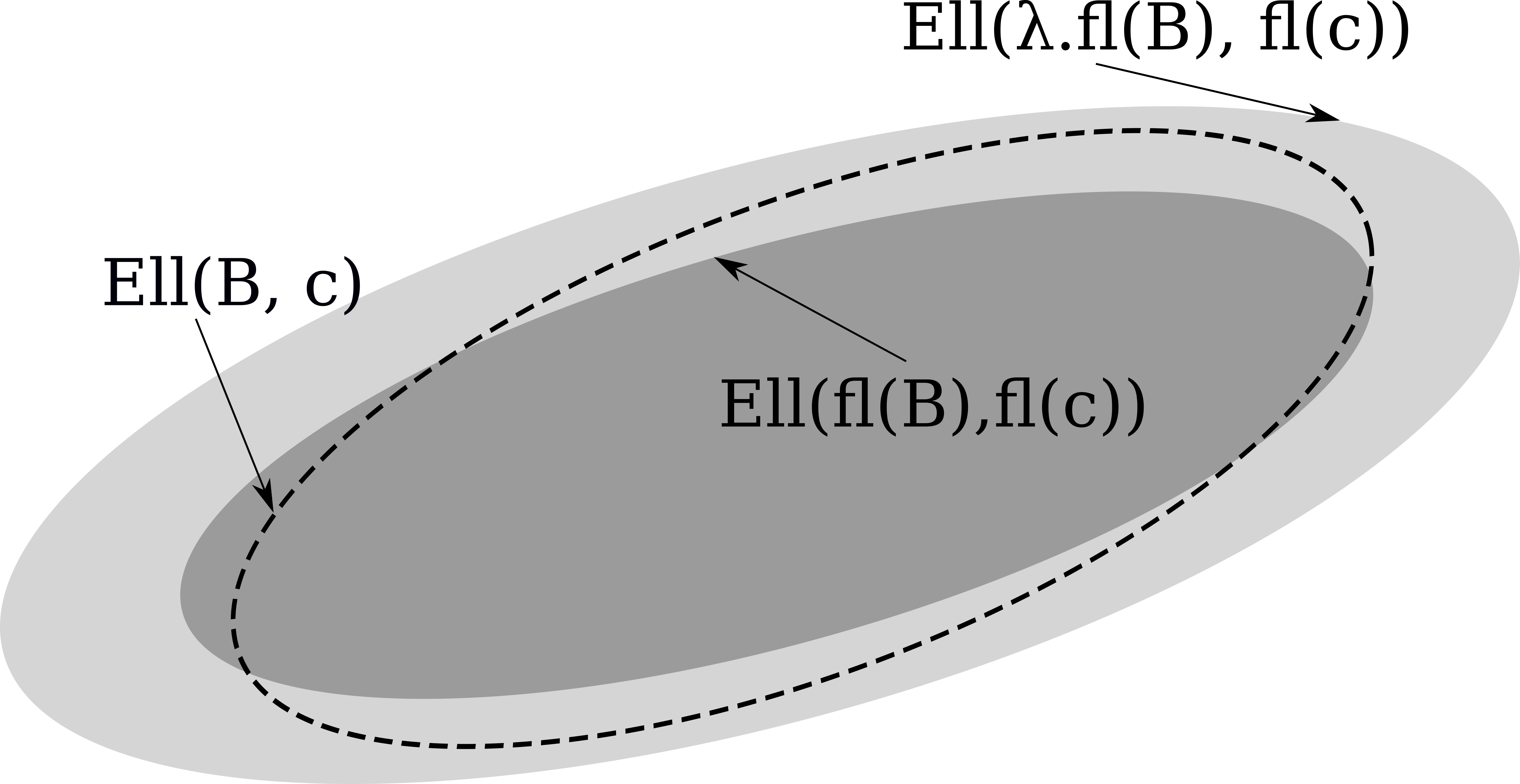}
\caption{Ellipsoid Widening \label{fig:ellWide}}
\end{center}
\vspace{-10pt}
\vspace{1pt}
\end{wrapfigure}

\noindent  In order to take into account the uncertainties on the variables due to 
floating-point rounding, the algorithm is modified to make it
more robust. Those uncertainties are first evaluated and a 
coefficient $\lambda$ is then computed. This coefficient represents how much the
ellipsoid $E_{k}$ is being widened at each iteration (see Fig.~\ref{fig:ellWide}).
Let us assume we have 
$B \in \mathbb{F}^{n \times n}$, $p \in \mathbb{F}^{n}$, $c \in \mathbb{F}^{n}$. 
We want to find $\lambda \geq 1\in \mathbb{R}$ such that:
\vspace{-5pt}
\begin{equation}
\label{eq:pb_formulation}
 \Ell \Big( B,c \Big) \subset 
 \Ell \Big( \lambda \cdot \fl(B),~\fl(c) \Big).
 \vspace{10pt}
\end{equation}
\noindent Before evaluating any of those rounding errors, Lemma 
\ref{lemma:Widening_Sufficient_Condition} is stated, which gives a sufficient condition 
for the coefficient $\lambda$ to have $\Ell \big( \lambda \cdot \fl(B),\fl(c) \big)$ 
covering $\Ell(B,c)$. If the calculations of $B$ and $c$ were perfect, using 
Lemma \ref{lemma:Widening_Sufficient_Condition}, $\lambda = 1$ would be a solution;
no correction is indeed necessary. 
\begin{mylemma}{[Widening - Sufficient Condition]}
\vspace{-5pt}
\[ \norm{ \fl\big(B\big)^{-1} B} + \norm{  \fl\big(B\big)^{-1}} \cdot \norm{c -  \fl(c)} \leq \lambda  \implies 
 \Ell \Big( B,c \Big) \subset \Ell \Big( \lambda \cdot \fl(B),~\fl(c) \Big).  \vspace{-5pt} \]
\label{lemma:Widening_Sufficient_Condition}
\end{mylemma}
\begin{po}
The starting hypothesis is:
$\norm{ \fl\big(B\big)^{-1} B} + \norm{  \fl\big(B\big)^{-1}} \cdot \norm{c -  \fl(c)} \leq \lambda
$. Using the fact that the two norm is a consistent norm, we have:
\vspace{-5pt}
\begin{equation}
\nonumber
\forall ~ u \in B_{1}(0)~,~ \norm{ \fl\big(B\big)^{-1} Bu}  
\leq \norm{ \fl\big(B\big)^{-1} B}
~~~\text{and}~~~~
\norm{  \fl\big(B\big)^{-1} \Big( c -  \fl(c) \Big) }  
 \leq \norm{  \fl\big(B\big)^{-1}} \cdot \norm{c -  \fl(c)} . \vspace{-5pt}
\end{equation} 
Therefore using these properties and the assumed starting property, we have:
\vspace{-5pt}
\[ \forall ~ u \in B_{1}(0)~,~ \norm{ \fl\big(B\big)^{-1} Bu}  +
 \norm{  \fl\big(B\big)^{-1} \cdot \Big( c -  \fl(c) \Big) } 
\leq \lambda . \vspace{-5pt} \]
Then, using the triangle Inequality ($\norm{x+y} \leq \norm{x} + \norm{y}$), 
we finally have:
\vspace{-5pt}
\[ \forall ~ u \in B_{1}(0)~,~ \norm{ \fl\big(B\big)^{-1} \cdot \Big ( Bu + c -  \fl(c) \Big) } 
\leq \lambda. \vspace{-5pt} \]
\noindent Reformulating the second part of the above statement (we are assuming that $\lambda >0$):
\vspace{-5pt}
\[ \forall ~ u_{1} \in B_{1}(0)~,~ z = B u_{1} + c \rightarrow  \exists ~ u_{2} \in B_{1}(0)~,~  \Big( \lambda \cdot \fl(B) \Big)^{-1} 
\cdot \Big(z-  \fl(c) \Big) =  u_{2}.  \vspace{-5pt} \]
Using the definition of Ellipsoids described in Eq.~\eqref{eq:ellipsoid_def}, 
one can write the equivalent statement:
\vspace{-5pt}
\[ \forall ~ z \in \Ell(B,c),  \exists ~ u_{2} \in B_{1}(0)~,~  \Big( \lambda \cdot \fl(B) \Big)^{-1} 
\cdot \Big(z-  \fl(c) \Big) =  u_{2}  . \vspace{-5pt} \]
Rearranging the equation implies:
\vspace{-5pt}
\[ \forall ~ z \in \Ell(B,c),  \exists ~ u_{2} \in B_{1}(0)~,~ 
z =  \lambda \cdot \fl(B) u_{2} +   \fl(c) . \vspace{-5pt} \]
Again, this is equivalent to:
\vspace{-5pt}
\[ \forall ~ z \in \Ell(B,c)  \rightarrow z  \in 
\Ell \Big( \lambda \cdot \fl(B), \fl(c) \Big). \vspace{-5pt} \]
which is the desired property:
\vspace{-5pt}
\[ \Ell \Big( B,c \Big) \subset \Ell \Big( \lambda \cdot \fl(B),~\fl(c) \Big). \]
\QEDB
\end{po}
\vspace{-5pt}
\noindent The floating-point errors $\Delta_{B}$, $\Delta_{B^{-1}}$ and $\Delta_{c}$ are defined as:
\vspace{-5pt}
\begin{equation}
\Delta_{B} = \fl(B) - B ~~~\text{,}~~~ 
\Delta_{c} = \fl(c) - c ~~~\text{,}~~~ 
\Delta_{B^{-1}} = (\fl(B))^{-1} - (B)^{-1}~~~\text{.}~~~ 
\vspace{-5pt}
\label{eq:delta_c_b}
\end{equation}
For now, it is assumed that after performing the floating-point analysis,
$\mathcal{E}_{B}$ and $\mathcal{E}_{c}$ have been found such that:
\vspace{-5pt}
\[ \abs{ (\Delta_{B})_{i,j} } \leq \mathcal{E}_{B} ~~~~\forall i,j \in [1,n], 
~~~~~~ \text{and} ~~~~~~
\abs{ (\Delta_{c})_{i} } \leq \mathcal{E}_{c}   ~~~~\forall i \in [1,n]. \vspace{-5pt} \]
%
From Lemma \ref{lemma:Widening_Sufficient_Condition}, one can see that the 
calculation of a widening coefficient $\lambda$ highly depends on the accuracy 
of the matrix $\fl(B)^{-1}$. Therefore, $\mathcal{E}_{B^{-1}}$ 
is also needed such that: $\abs{ (\Delta_{B^{-1}})_{i,j} } \leq 
\mathcal{E}_{B^{-1}} ~~~\forall i,j \in [1,n]$. \\
The quantity $(B)^{-1}$ is not used explicitly in the
algorithm and its floating-point error could not be evaluated
by numerically analyzing the method.
Instead, perturbation matrix theory~\cite{Ghaoui2002} 
and Theorem \ref{th:inverse_error} will be used, which gives a lower bound 
on $\mathcal{E}_{B^{-1}}$ given $\mathcal{E}_{B}$, the norm of $B$ 
and its condition number. The result is stated as follows.
%
\begin{mylemma}{[Widening - Analytical Sufficient Condition]} \\
\vspace{-5pt}
\[ 1+ \frac{k(B)}{\norm{B}} \sqrt{n} \cdot \bigg( \sqrt{n} \cdot 
k(B) \mathcal{E}_{B} + \mathcal{E}_{c} + \frac{k(B)}{\norm{B}} n \mathcal{E}_{B} \mathcal{E}_{c}  \ \bigg) \leq \lambda 
 \implies  \Ell \Big( B,c \Big) \subset \Ell \Big( \lambda \cdot \fl(B),~\fl(c) \Big). \]
\vspace{-10pt}
\label{lemma:Widening_Analytic_Sufficient_Condition}
\end{mylemma}
\begin{po}
To prove this lemma, $\norm{\Delta_{B^{-1}}}$ is first evaluated.
Using Eq.~\eqref{eq:inv_pert}:
\[ \norm{\Delta_{B^{-1}}} \leq k(B) \frac{\norm{B^{-1}}}{\norm{B}} \norm{\Delta_{B}} =  \frac{k^{2}(B)}{\norm{B}^{2}} \norm{\Delta_{B}}. \]
But, $ \norm{\Delta_{B}} \leq \norm{\Delta_{B}}_{F} \leq n \mathcal{E}_{B}$, which implies: 
$\norm{\Delta_{B^{-1}}} \leq \frac{k^{2}(B)}{\norm{B}^{2}} n \mathcal{E}_{B}$.
The three constants $I, J$ and $K$ are now defined:
$I = \norm{\fl\big(B\big)^{-1} B}$, 
$J = \norm{\fl\big(B\big)^{-1}}$ and 
$K =  \norm{c -  \fl(c)} $. \\
The next step consists of computing an upper bound for each of those constants.
\vspace{-10pt}
\[ I =  \norm{ I_{n} + \Delta_{B^{-1}} B }  \implies I \leq  
\norm{I_{n}} + \norm{ \Delta_{B^{-1}}} \norm{B} =
1 +  \frac{k^{2}(B)}{\norm{B}} n \mathcal{E}_{B}, \vspace{-5pt} \]
\vspace{-10pt}
\[ J \leq   \norm{(B)^{-1}}  + \norm{\Delta_{B^{-1}}} = 
\frac{k(B)}{\norm{B}} \cdot \Big(1+ \frac{k(B)}{\norm{B}} n \mathcal{E}_{B} \Big) , \vspace{-5pt} \]
\vspace{-5pt}
\[ K \leq \sqrt{n} \mathcal{E}_{c} .  \vspace{-5pt} \]
So, if
\vspace{-5pt}
\[   1+ \frac{k(B)}{\norm{B}} \sqrt{n} \cdot \bigg( \sqrt{n} \cdot 
k(B) \mathcal{E}_{B} + \mathcal{E}_{c} + \frac{k(B)}{\norm{B}} n \mathcal{E}_{B} \mathcal{E}_{c}  \ \bigg) \leq \lambda \implies    I + J \cdot K \leq \lambda ,   \]
then
\[    \norm{ \fl\big(B\big)^{-1} B} +  \norm{  \fl\big(B\big)^{-1}} \cdot  \norm{c -  \fl(c)}  	\leq \lambda .   \]
Using the result of Lemma \ref{lemma:Widening_Sufficient_Condition},
one can conclude on the inclusion property~\eqref{eq:pb_formulation}.
\QEDB
\end{po}
\noindent Thus, due to Lemma~\ref{lemma:Widening_Analytic_Sufficient_Condition}, following the floating-point analysis of
the algorithm, a coefficient $\lambda$ such that 
Eq.~\eqref{eq:pb_formulation} is valid can now be computed.
After finding such a $\lambda$, using over-approximation schemes,
we consider how the algorithm's convergence changes. Because the method's proof
lies with the fact that the final ellipsoid has a small enough volume,
this correction has an impact on the guaranteed number of iterations. Lemma
\ref{th:cv_lambda} addresses those issues.
\begin{mylemma}{[Convergent Widening Coefficient]}
Let $n \in \mathbb{N}, n \geq 2$. \\
The algorithm implementing the widened ellipsoids, with coefficient $\lambda$ converges if:
\vspace{-5pt}
\begin{equation}
\lambda < \exp \big(  1/ (n(n+1))  \big).
\label{eq:cv_lambda}
\vspace{-5pt}
\end{equation}
In that case, if $N$ denotes the original number of iteration needed, 
the algorithm implementing the widened ellipsoids will require:
\vspace{-5pt}
\begin{equation}
N_{\lambda} = N / \big( 1-n(n+1) \cdot \log(\lambda) \big) ~~ \text{ iterations}
\label{eq:it_floats}
\end{equation}
\vspace{-5pt}
\label{th:cv_lambda}
\end{mylemma}
\vspace{-5pt}
\begin{po}
We recall that the Ellipsoid algorithm implementing the widened ellipsoids, with
coefficient $\lambda$ converges if:
\vspace{-5pt}
\[ \Vol \big( \lambda \cdot \Ell(B_{k+1},c_{k+1}) \big)  <  \Vol \big( \Ell(B_{k},c_{k}) \big).  \vspace{-10pt}\]
When using the Ellipsoid algorithm update process (with or without condition number correction), we have:
\vspace{-5pt}
\[ \Vol \big( \Ell(B_{k+1},c_{k+1}) \big)  \leq \gamma \cdot \Vol \big( \Ell(B_{k},c_{k}) \big)  \vspace{-10pt} \] 
With $\gamma = \exp(-1/(2(n+1)))$, and Eq.~\eqref{eq:volume_ellipsoid}, we know that:
\vspace{-5pt}
 \[ \Vol \big( \Ell(\lambda \cdot B_{k+1},c_{k+1}) \big) = \lambda^{n/2} \cdot \Vol \big( \Ell( B_{k+1},c_{k+1}) \big)   \vspace{-10pt} \]
Therefore, the algorithm implementing the widened ellipsoids, with widening coefficient $\lambda$ will converge if:
\vspace{-5pt}
\[ \lambda^{n/2} \cdot \gamma < 1  ~~~~\text{which is equivalent to}~~~~ \lambda  < \exp \big(1/(n(n+1)) \big) \vspace{-20pt}  \]
\QEDB 
\end{po}
\vspace{-15pt}
\begin{po}
\noindent The second statement of Lemma~\ref{th:cv_lambda} is now proved, which shows 
how to compute the updated number of iterations for the algorithm implementing widened ellipsoids.
We introduce $B^{\prime}_{k+1} = \lambda B_{k+1}$.
Therefore using Eq.~\eqref{eq:volume_ellipsoid}, and
$\Vol(E^{\prime}_{k+1}) = \lambda^{n/2} \cdot \Vol(E_{k+1}) $
, we get:
\vspace{-10pt}
\[ \gamma_{\lambda} = \frac{\Vol(E^{\prime}_{k+1} ) }{\Vol (E_{k}) } = 
\frac{\Vol(E^{\prime}_{k+1} ) }{ \Vol (E_{k+1}) } \frac{\Vol(E_{k+1}) }{ \Vol (E_{k}) }  =
\lambda^{n/2} \cdot \gamma. \vspace{-10pt} \]
To end up at the final step with an ellipsoid of the same volume, we
need: $\Vol(E_{o}) \cdot \gamma_{\lambda}^{N_{\lambda}} =  \Vol(E_{o}) \cdot \gamma^{N}$. Which implies
\vspace{-5pt}
\[ N_{\lambda}  \big( \log(\gamma) + n/2 \cdot \log (\lambda )  \big) =   N \log (\gamma).  \vspace{-5pt}  \]
Replacing $\gamma$ by its value, using property \ref{prop:volume_ratio} from Section~\ref{sec:EllMethod}:
\vspace{-5pt}
\[ N_{\lambda}  \Big( n/2 \cdot \log (\lambda ) -1/(2(n+1))  \Big) = 
 -N /(2(n+1))  \vspace{-5pt}  \]
And thus, we arrive at the formula:
\vspace{-10pt}
\[ N_{\lambda}  =  N / \{1 - n(n+1) \log (\lambda ) \}. \vspace{-20pt} \]
\QEDB
\end{po}
\vspace{-10pt}
\subsection{Computing $\mathcal{E}_{B}$ and $\mathcal{E}_{c}$}
\label{sec:flp_epsilon}
\noindent This section introduces the evaluation of the floating-point errors taking place 
when performing the update formulas \eqref{eq:updateCenter} and \eqref{eq:updateShape}
(represented by  $\mathcal{E}_{c}$ and $\mathcal{E}_{B}$).
For this, numerical properties for basic operations appearing in the algorithm
are first presented. \\
\emph{\underline{Rounding of a Real.}} Let $z \in \mathbb{R}$,
$ ~~~~ \tilde{z} = \fl(z) = z + \delta + \eta ~~~~~ \text{with} ~ |\delta|< \textbf{u}
~ \text{and} ~ |\eta|< \textbf{eta}/2$ \\
\emph{\underline{Product and Addition of Floating-Points.}} Let $a,b \in \mathbb{F}$.
\vspace{-10pt}
\[ \fl(a + b) ~=~ (a + b)(1+\epsilon_{1})  ~~~~~~~ 
\quad \quad \quad \quad \quad \quad \quad \text{with} \quad 
\abs{\epsilon_{1}} < \textbf{u}  \vspace{-20pt} \]
\vspace{-5pt} 
\[ \fl(a \times b) ~=~ (a \times b)(1+\epsilon_{2}) + \eta_{2}  ~~~~~~~ 
\text{with} \quad  \abs{\epsilon_{2}} < \textbf{u},~\abs{\eta_{2}} < \textbf{eta}  
~~\text{and}~~ \epsilon_{2} \cdot \eta_{2} = 0  \vspace{-5pt}  \]
\emph{\underline{Reals-Floats Product.}} Let $z \in \mathbb{R}$ and $a \in \mathbb{F}$, ~~~ $ | \fl \big( \fl(z) \cdot a \big) - z \cdot a |  \leq      |z||a| \cdot  \textbf{u} + |a| \cdot	2 \textbf{u} (1+\textbf{u})$ 
\clearpage
\newpage
\noindent \emph{\underline{Scalar Product.}} Let $a,b \in \mathbb{F}^{n}$. We define,
$\langle a,b \rangle ~=~ \sum_{i=1}^{n} a_{i} b_{i}$ and 
$ | a,b | ~=~ \sum_{i=1}^{n} |a_{i} b_{i}|$. 
We have then: 
\vspace{-7pt}
\[ \left| \fl \langle a,b \rangle - \langle a,b \rangle  \right| \leq  A_{n} | a,b |   + \Gamma_{n} \]
\vspace{-25pt}
\begin{equation}
\nonumber
\text{With:} \quad A_{n} = n \cdot \textbf{u} / (1-n \cdot \textbf{u}) \quad  \text{and} \quad \Gamma_{n} = 
A_{2n} \cdot \textbf{eta} / \textbf{u}  
\vspace{-5pt}
\end{equation}
\vspace{-10pt}
\paragraph*{Rounding Error on $c$.}
Knowing how the errors are being propagated through elementary transformations, 
the goal is to compute the error for a transformation similar to the 
vector $c$'s update. For each component of $c$, we have:
\vspace{-5pt}
\begin{equation}
\nonumber
c_{i} =  c_{i} - \Big( 1 / \{ n+1 \} \Big) \cdot  \langle \text{Row}_{i}(B) , p \rangle.
\vspace{-5pt}
\end{equation}
\noindent Therefore, the operation performed, in floating-point arithmetic is:
\vspace{-5pt}
\begin{equation}
\fl \big( c + \fl \big( \fl(z) \cdot \fl\langle a,b \rangle \big) 
~~~~~~~~~~\text{with:}~~ a,b \in \mathbb{F}^{n}, ~ c \in \mathbb{F} ~\text{and} ~  z \in \mathbb{R}.
\label{eq:floating_transfomation_x}
\vspace{-5pt}
\end{equation}
Using this and neglecting all terms in \textbf{eta} and powers of $\textbf{u}$ greater than two, we get:
\vspace{-10pt}
\begin{equation}
\mathcal{E}_{c}   \leq u \cdot \Big( \big( 16n^2 + 16n + 3 \big) \cdot \norm{B} + \norm{c}  \Big).
\label{eq:flp_eps_c}
\vspace{-5pt}
\end{equation}
\paragraph*{Error on $B$.} Similarly, for each component of $B$, the operation below is performed:
\vspace{-5pt}
\begin{equation}
\nonumber 
B_{i,j} =  \alpha \cdot B_{i,j}  + \beta  \cdot 
\langle \text{Row}_{i}(B) , p \rangle \cdot p_{j}.
\vspace{-5pt}
\end{equation}
In floating-point arithmetic, this last equation can be put into form: 
\vspace{-5pt}
\begin{equation}
\nonumber
\fl \bigg( \fl \Big( \fl(z_{1}) \cdot d \Big)  + 
\fl \Big(  \fl(z_{2}) \cdot  \fl  \big(  \fl  \langle a,b \rangle \cdot c \big) \Big)  \bigg) .
\label{eq:floating_transfomation_B}
\vspace{-10pt}
\end{equation}
Similarly, propagating the errors using elementary transformations implies that:
\vspace{-5pt}
\begin{equation}
\mathcal{E}_{B}   \leq u \cdot \norm{B} \cdot \Bigg( \big( n^2 / \{ 1-nu \} +2 \big) |\beta| + n + 2 |\alpha| + 1  \Bigg).
\vspace{-5pt}
\label{eq:flp_eps_B}
\end{equation}
%

%% file: experiment.tex
\section{Automatic Code Generation and Examples}
\label{sec:experiment}
\subsection{Credible Autocoding}
\label{sec:craves}
\begin{figure}[ht!]
\centering
\includegraphics[scale=0.8]{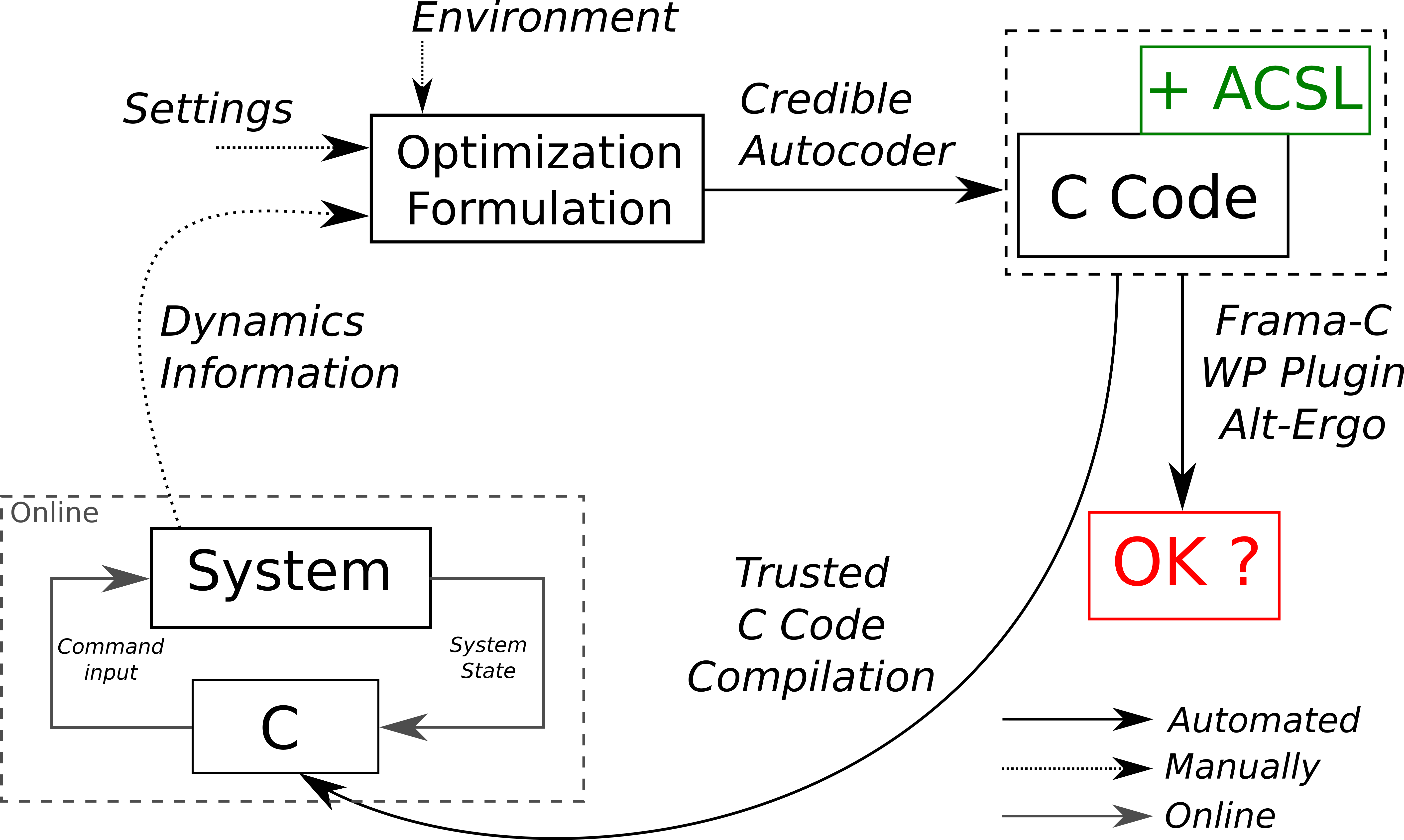}
\caption{Credible Autocoding Framework}
\label{fig:craves}
\end{figure}

\noindent  Credible autocoding, is a process by which an implementation of a certain input model
in a given programming language is being generated along with formally verifiable
evidence that the output source code correctly implements the input model.
The goal of the work presented in this article is to automatically 
generate, formally verifiable C code implementations of a given receding horizon
controller. Thus, an autocoder that we call a \say{Credible Autocoder} 
(see Fig.~\ref{fig:craves} ) has been built 
that generates an ACSL annotated C code implementation of a given MPC controller. 
This autocoder takes as an input a formulation of a MPC controller written by the 
user in a text file. Once the output code is generated, it can be checked using 
the software analyzer Frama-c and the plugin WP. 
If the verification terminates positively, the code correctly implements
the wanted receding horizon control.
The controller can then be compiled and the binary file embedded 
in a feedback control loop.
The specification and the requirements are automatically generated from
the input text file written by the user.
The verification taking place in this work is applied to 
one high-level requirement (of the MPC solver). 
The other high-level requirements such as those involving control-related 
issues are not examined.  
\begin{wrapfigure}{r}{0.50\textwidth} 
\vspace{-25pt}
\begin{center}
\includegraphics[width=0.95\linewidth]{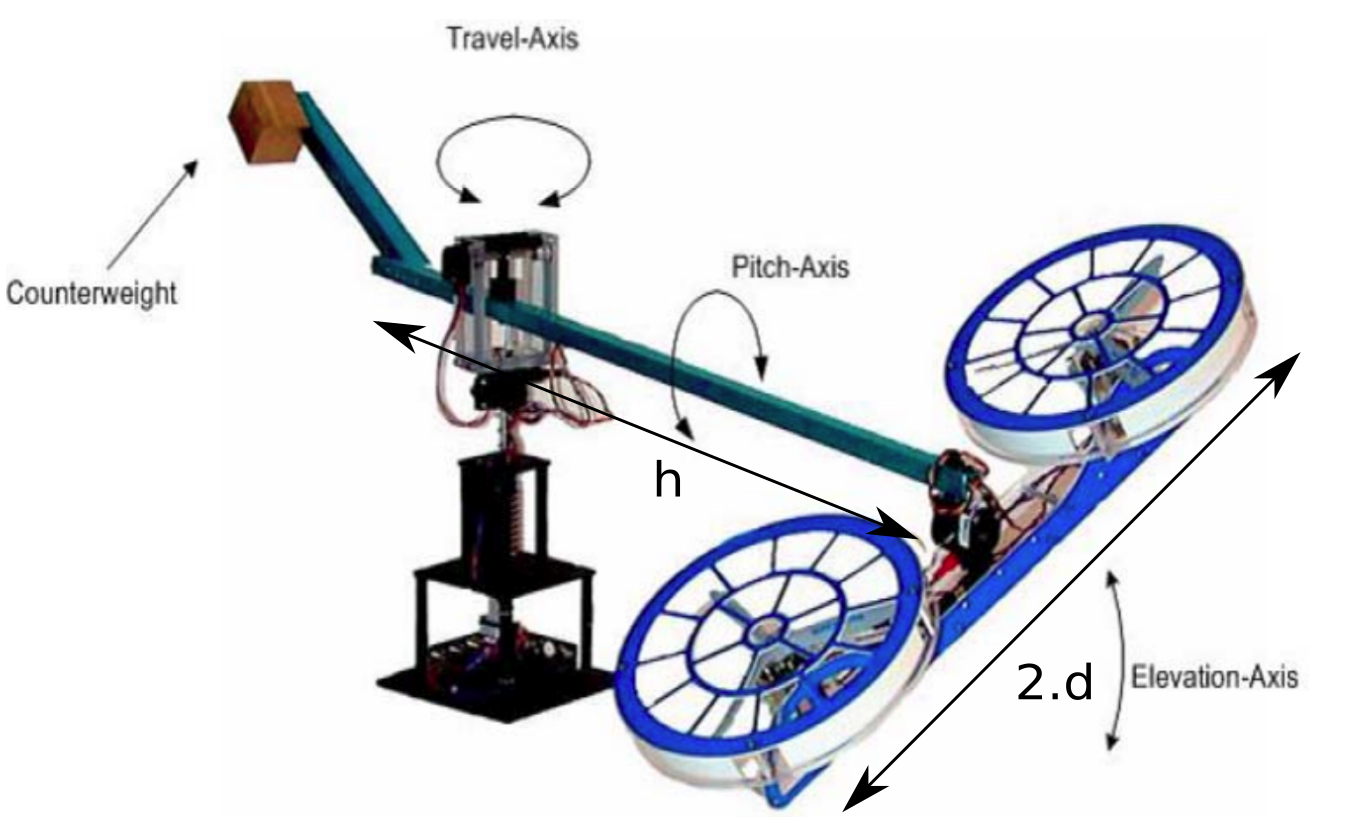}
\caption{Quanser -- 3 DOF Helicopter}
\label{fig:quanser}
\end{center}
\vspace{-40pt}
\end{wrapfigure} 
Nevertheless, given that the input text file is written in a high-level 
language, specifically designed for MPC formulations, 
it is relatively simple to read. The use of autocoders for 
automated MPC code generation makes the formulation easier to 
check and add traceability to the algorithms.
\subsection{Example: the three degree-of-freedom helicopter}
\begin{figure}[ht!]
\centering
\begin{subfigure}{.5\textwidth}
\centering
\includegraphics[scale=0.6]{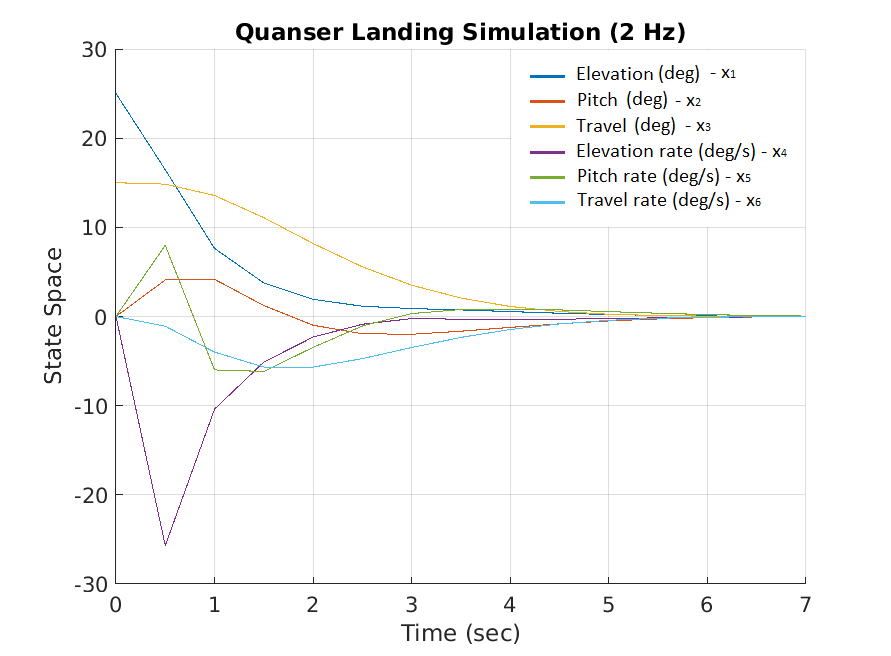}
\caption{State Vector versus time}
\label{fig:state_x_quanser_landing}
\end{subfigure}%
\begin{subfigure}{.5\textwidth}
\centering
\includegraphics[width=1.10\linewidth]{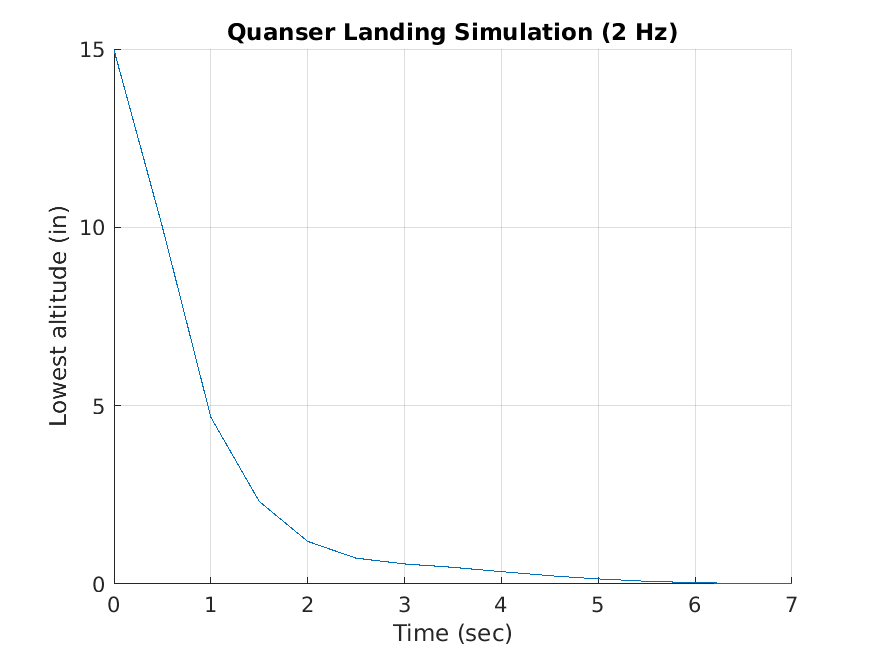}
\caption{Lowest Altitude versus time}
\label{fig:altitude_quanser_landing}
\end{subfigure}
\caption{Simulation Data \label{fig:data_experiment}}
\end{figure}

\noindent The three degree-of-freedom (3 DOF) helicopter shown in Fig.~\ref{fig:quanser}
was used to illustrate the framework developed in this article.
The vector state of the system collects the 3 axis angles and rates and it
is denoted by $x=[\theta~\psi~\phi~\dot{\theta}~\dot{\psi}~\dot{\phi}]$. The 
inputs are the voltages of the front and back DC motors. Further information about the 
3 DOF helicopter can be found in~\cite{quanser}. Given an inner 
feedback controller and a discretization step of $T=0.5~sec$, the resulting 
system is a stable linear system.
The problem we are trying to solve is 
a landing of the 3 DOF helicopter. Starting with an angle of $25~deg$ in elevation,
$15~deg$ in travel (see Fig.~\ref{fig:quanser} for axis) 
and all the other states being zero, the objective is
to design a controller that can drive
the system back to the origin while avoiding the ground.
The ground is the area below $\theta = 0$. 
Thus, the constraint for enforcing ground avoidance is a formula on the elevation and the pitch 
angles and can be formulated as: $h \sin(\theta) \pm d \sin(\phi) >= 0$. 
By linearizing this constraint over small angles, one can obtain linear inequalities of the 
form: $A_{obs}x \leq b_{obs}$. We want to implement the following MPC controller.
\begin{equation}
\setlength{\jot}{-2pt}
\label{opt3}
\begin{aligned}
&  \underset{X = [\textbf{x}, \textbf{u}]}{\text{minimize}}
& & \sum_{k=1}^{H} ~~ \norm{x_{k}} \\
& & & x_{k+1} = A x_{k} + B u_{k} ~,~ k=1..H-1 \\
& & & \norm{u_{k}} \leq 60  ~,~ k=1..H-1 \\
& & & 0 \leq x_{k}(1)  ~~\text{and}~~ A_{obs} x_{k} \leq b_{obs}  ~,~ k=2..H \\
& & &  x_{1} = \hat{x}_o   \\
\end{aligned}
\end{equation}
Assuming that $\norm{\hat{x}_o} \leq 27$, we found, using the method developed in 
Section~\ref{sec:online} a radius $8.0612$ (running an off-line optimization 
problem that finds the largest ball inside $P_{min}$). Similarly, $R = 322$,
was obtained using Eq.~\eqref{eq:R_mpc}.
From the problem formulation, one can see that the norms of the successive $x$'s along
the trajectory constructed are supposed to be minimized. Given a starting point, an
upper bound on the objective function over the feasible set can be computed.
The objective function is maximal when $x_o$ has the largest norm, and when the system 
stays at this point throughout the trajectory. The constant $V$ can therefore be computed as:
\vspace{-5pt}
\begin{equation}
V = H  \cdot \norm{x_o} ~ \leq ~ H \cdot 27 = 6 \times 27 = 162.
\vspace{-5pt}
\end{equation}
Following those calculations, a number of step of $N = 5528 $
is found.
In order to control floating-point errors, a widening coefficient 
$\lambda$ can be constructed. Performing the steps described in Section~\ref{sec:flp},
using double precision floating-points and an accuracy of $\epsilon = 0.25$
result in:
\begin{equation}
\nonumber
\lambda = 1.000695409372118.
\end{equation}
As it was explained in Section~\ref{sec:flp}, the ellispoid widening
increased the number of iterations needed for the convergence to
$N_{\lambda} = 6817$.
The simulation was ran on a Intel Core i5-3450 CPU @ 3.10GHz $\times$ 4 processor 
and the running time was approximately $0.2~sec$ for a single point.
The results of the simulation can be found in
Fig.~\ref{fig:state_x_quanser_landing} and~\ref{fig:altitude_quanser_landing}.
Figure~\ref{fig:state_x_quanser_landing} presents
the closed-loop response for the state vector $x$ and 
Fig.~\ref{fig:altitude_quanser_landing} shows the lowest altitude point
with time for the same simulation.
The text file presented in Fig.~\ref{fig:autocoder_input} has been used to 
generate the C code in order to perform the simulation. The whole simulation was
executed using the autocoded C code and a Simulink model. The full autocoder source code, 
input file (Fig.~\ref{fig:autocoder_input}), Simulink model 
and a user guide for the autocoder can be found online\footnote{The source code for 
the autocoder is available at: \url{https://cavale.enseeiht.fr/quanser_mpc/}}. 

\begin{figure}[ht!]
\centering
\begin{genemoListing}{Input File to the Autocoder}
Input
xo(6)
Output
u(:,1)
Constants
H = 6; M = H-1; l = 90; r = 40;
A = [0.7101    0.0000   -0.0000    0.2331    0.0000    0.0000;
     0.0000    0.2105    0.4023    0.0000    0.0977    0.7390;
    -0.0000   -0.1272    0.9846   -0.0000   -0.0134    0.4733;
    -0.8721    0.0000   -0.0000    0.0724    0.0000    0.0000;
    -0.0000   -2.0777    0.7830    0.0000   -0.2674    1.6711;
    -0.0000   -0.4224   -0.1072   -0.0000   -0.0618    0.8109];
B = [0.2899    0.0000;    -0.0000   -0.4023; 0.0000    0.0154;
     0.8721    0.0000;     0.0000   -0.7830; 0.0000    0.1072];
Aobs = [-l  -r  0  0  0  0; -l  r  0  0  0  0]; bosbt = [0;0];
Variables
x(6,H) u(2,M)
Minimize
sum( || x(:,k) || , k = 1..H )
SubjectTo
constraint1: x(:,1) = xo;
constraint2: x(:,k+1) = A*x(:,k) + B*u(:,k)	 ,k=1..H-1;
constraint3: -30 <= u(1,k)	                 ,k=1..H-1;
constraint4: u(1,k) <= 30                    ,k=1..H-1;
constraint5: -30 <= u(2,k)	                 ,k=1..H-1;
constraint6: u(2,k) <= 30                    ,k=1..H-1;
constraint8:  0 <= x(1,k)                    ,k=2..H;
constraint9: -40 <= x(2,k)	                 ,k=2..H;
constraint10: x(2,k) <= 40                   ,k=2..H;
constraint11: Aobs*x(:, k) <= bosbt	         ,k=2..H;
Information
r = 8.06; R = 322; V = 162; eps = 0.25; lambda = 1.000695409372118;
\end{genemoListing}
\caption{3 DOF Helicopter Landing Problem: Autocoder Input File}
\label{fig:autocoder_input}
\end{figure}

%% file: conclusion.tex
\section{Conclusion}
\label{sec:conclusion}
\noindent In this article, we presented a formal framework for the automatic 
generation and verification of optimization code for solving 
second-order cone programs. We focused on the ellipsoid method due to 
its good numerical characteristics.   
We built a framework capable of compiling the high-level requirements 
of online receding horizon solvers into C code programs which can then be
automatically verified using existing formal methods tools.
The credible autocoding framework developed is targeting a certain type of
convex optimization problems and the high-level requirements formalized are
appropriately chosen. However, if additional high-level requirements
are needed, some manual
formalization are needed. Although high-level requirements
formalization can be complex, 
the struggle during this task is to formalize the low-level mathematical
types and predicates needed.
Hence, this task being already done, the same mathematical foundations can be used and 
the formalization of additional high-level requirements 
within the credible autocoder is highly simplified.
\newline A numerical analysis of the method has been presented, showing 
how to propagate the errors due to floating-point calculations 
through the operations performed by the program. A modified version
of the algorithm was presented, allowing us to compute \say{reasonable}
{\em a priori} bounds on floating-points errors. 
However, the numerical analysis performed remains for now purely manual.
Its correctness depends on the exactitude of equations obtained manually
and no verification tools was used for this part. Future work shall
include the use of formal methods to validate the numerical analysis.